\documentclass[preprint,12pt]{elsarticle}

\usepackage{graphicx}
\usepackage{multirow}
\usepackage{amsmath,amssymb,amsfonts,mathtools}
\usepackage{amsthm}
\usepackage{mathrsfs}
\usepackage[title]{appendix}
\usepackage[table]{xcolor}
\usepackage{textcomp}
\usepackage{booktabs}
\usepackage{array}
\usepackage{url}
\usepackage{bm}
\usepackage{threeparttable}
\usepackage{adjustbox}
\usepackage{wasysym}
\usepackage{hyperref}
\hypersetup{hidelinks}

\usepackage{pifont}
\usepackage{colortbl}   
\newcommand{\cmark}{\textcolor{green!60!black}{\ding{51}}}
\newcommand{\xmark}{\textcolor{red}{\ding{55}}}
\usepackage{float}
\usepackage{makecell}
\usepackage{placeins}

\definecolor{posgreen}{RGB}{0,110,70}
\definecolor{rowours}{RGB}{232,242,252}
\definecolor{rowbase}{RGB}{248,248,246}

\newcommand{\pos}[1]{\textcolor{posgreen}{\textbf{+#1}}}
\newcommand{\nd}{n.a.}

\journal{Information Processing \& Management}

\begin{document}

\begin{frontmatter}

\title{SegRAG: Retrieval Augmented Spatial Prompting for Open Vocabulary Semantic Segmentation\tnoteref{license}}

\tnotetext[license]{\textcopyright{} 2026. This manuscript version is made available under the CC BY-NC-ND 4.0 license: \url{https://creativecommons.org/licenses/by-nc-nd/4.0/}. The Version of Record is available at \url{https://doi.org/10.1016/j.ipm.2026.105190}.}

\author[ku]{Abderrahmene Boudiaf\corref{cor1}}
\ead{100058322@ku.ac.ae}

\author[ku]{Irfan Hussain}
\ead{irfan.hussain@ku.ac.ae}

\author[ku]{Sajid Javed}
\ead{sajid.javed@ku.ac.ae}

\cortext[cor1]{Corresponding author.\\
  \textit{Postal address:} Department of Electrical Engineering \&
  Computer Science, Khalifa University of Science and Technology,
  P.O.\ Box 127788, Abu Dhabi, United Arab Emirates.}

\affiliation[ku]{
  organization={Department of Electrical Engineering \& Computer Science,
                Khalifa University of Science and Technology},
  addressline={P.O.\ Box 127788},
  city={Abu Dhabi},
  country={United Arab Emirates}
}

\begin{abstract}
Frozen segmentation foundation models often fail when the target class appears in a form that is weakly represented during pretraining. To address this problem, we introduce SegRAG, a retrieval-augmented inference-time spatial prompting pipeline for open-vocabulary semantic segmentation that uses frozen models without updating their weights. SegRAG builds a compact class-indexed memory from annotated references. When multiple references are available, Intra-Class Cohesion Distillation (ICCD) filters DINOv3 patch descriptors by cross-image foreground agreement. With one reference, foreground descriptors are retained directly without ICCD. Topographic Similarity Grounding (TSG) then turns high-similarity query regions into point prompts for SAM\,3. In the matched five-shot comparison, SegRAG exceeds recent exemplar- and retrieval-based baselines on ADE20K-150, Cityscapes, and PC-59. It also improves over the SAM\,3 text-only baseline by $1.15$ to $3.92$ mean Intersection over Union (mIoU) points. In the up-to-30-shot AgML agricultural domain-transfer evaluation, SegRAG raises mIoU from $25.27$ to $59.24$. It also recovers text-only failures, including cauliflower from $0.00$ to $95.36$ IoU and sugarbeet weed from $0.00$ to $80.22$ IoU. Controlled ablations show complementary contributions from ICCD, TSG point selection, and joint text-and-point prompting. SegRAG therefore formulates segmentation adaptation as an information organization and retrieval problem by maintaining annotated visual knowledge as an external, class-indexed memory that can guide a frozen segmentation model without weight updates. Code: \url{https://github.com/boudiafA/SegRAG}.
\end{abstract}


\end{frontmatter}

\begin{highlights}
  \item SegRAG adapts frozen SAM\,3 through inference-time spatial prompting.
  \item ICCD filters multi-reference class banks by cross-image retrieval reliability.
  \item TSG turns retrieved patches into spatial point prompts for SAM\,3.
  \item Five-shot SegRAG exceeds four matched-data baselines on three benchmarks.
  \item SegRAG uses 28.8 AgML references per class and raises mIoU from 25.27 to 59.24.
\end{highlights}

\section{Introduction}
\label{sec:introduction}

Foundation models have changed semantic segmentation by allowing users to request masks through text, points, boxes, or visual examples rather than through a fixed task-specific output layer. This shift is valuable for information processing systems because the set of relevant visual entities often changes after deployment. A farm monitoring system may need a new crop growth stage, a medical archive may contain a rare imaging pattern, and an industrial inspection system may encounter a previously unseen defect. In these settings, retraining a large model is slow, costly, and sometimes impossible because only a small number of annotated examples are available.

\begin{figure}[!t]
  \centering
  \includegraphics[width=\linewidth]{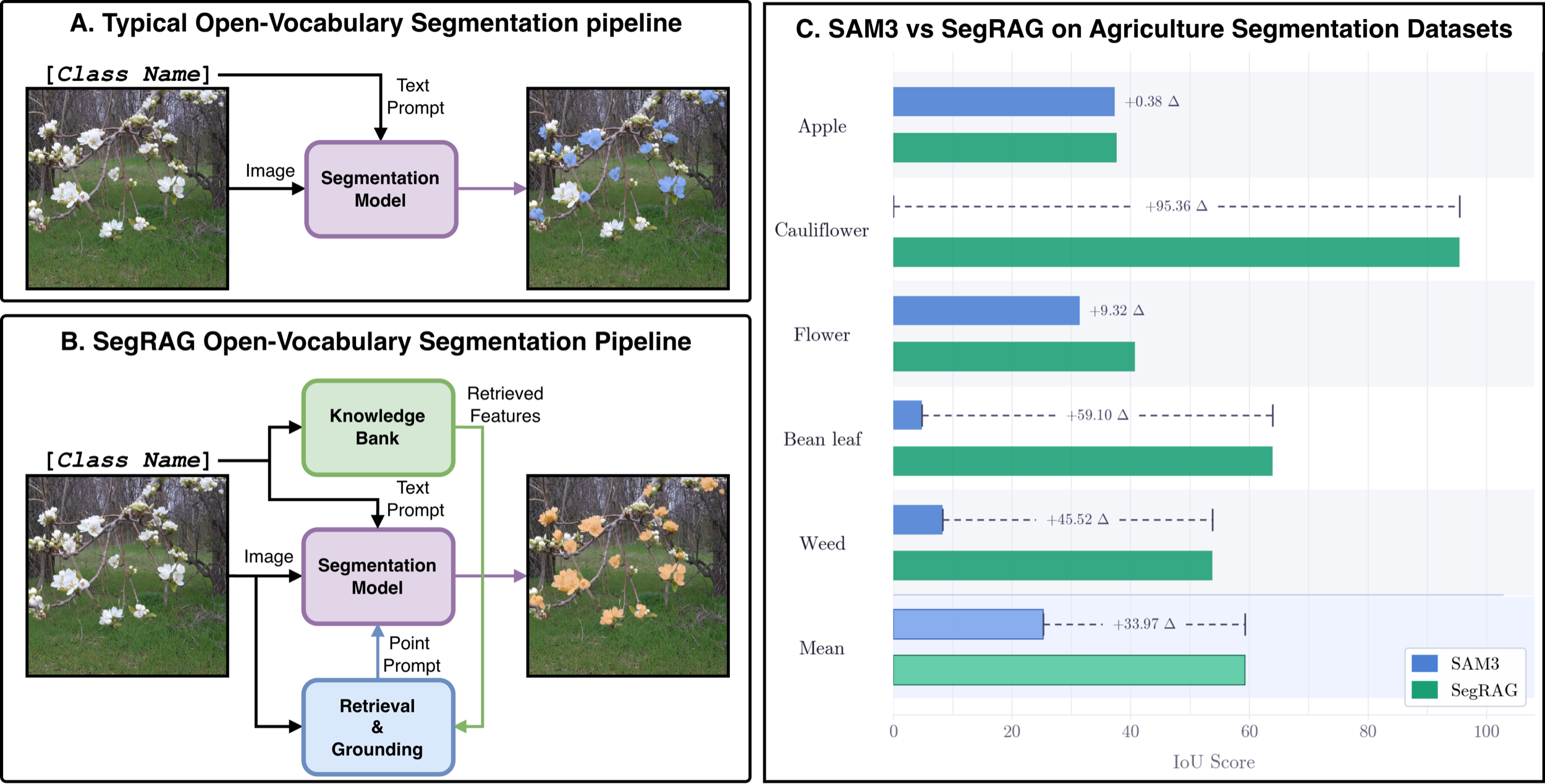}
  \caption{SegRAG overview. Text-only segmentation supplies the class name but lacks spatial evidence. SegRAG adds a class-indexed memory of annotated visual references. Retrieved descriptors are matched to the query image, converted into point prompts, and passed with the class name to SAM\,3. The agricultural examples show that the largest gains occur when the target appearance departs from common web imagery.}
  \label{fig:pipeline_overview}
\end{figure}

The limitation is not only a matter of vocabulary size. A class name gives the model semantic intent, but it does not tell the decoder where domain-specific evidence appears in the query image. SAM\,3 formalizes promptable concept segmentation and provides a strong text-guided baseline, yet its knowledge remains stored in model weights learned from broad visual data~\cite{carion2025sam3}. When a category appears in an unfamiliar viewpoint, growth stage, sensor modality, or texture regime, the text representation can be poorly aligned with the local image evidence. The same problem is visible in long-tailed benchmarks such as LVIS~\cite{gupta2019lvis} and in agricultural images, where early crop and weed morphology differs sharply from common web depictions. The resulting information-management problem is how to organize and retrieve limited domain-specific visual evidence for a frozen model.

Retrieval-augmented generation offers a natural response to this limitation. Instead of relying only on parametric knowledge, a frozen model can be grounded by non-parametric evidence retrieved at inference time~\cite{lewis2020rag}. This paper applies that principle to prompt formation for segmentation. We propose \textbf{SegRAG}, a retrieval-augmented inference-time spatial prompting pipeline that places a curated, class-indexed visual memory between annotated references and a frozen SAM\,3 decoder. SegRAG does not introduce a new segmentation architecture or update DINOv3 or SAM\,3 weights. The memory is built from DINOv3 patch descriptors because these descriptors are well suited for dense visual correspondence~\cite{simeoni2025dinov3}. During inference, retrieved descriptors are localized in the query image and converted into point prompts. SAM\,3 then receives these spatial prompts together with the class name in a single grounding call.

Figure~\ref{fig:pipeline_overview} summarizes the idea. The method is designed for settings where a user can supply a small set of annotated references but cannot fine-tune the underlying foundation model. The empirical results show that this design helps both standard open-vocabulary benchmarks and a domain-transfer case in agriculture. One- and five-shot SegRAG is evaluated on four public benchmarks. Under matched five-shot data and evaluation protocols on ADE20K-150, Cityscapes, and PC-59, SegRAG exceeds GF-SAM, FSSDINO, IPSeg, and Matcher. Across all four benchmarks, five-shot SegRAG improves the directly controlled SAM\,3 text-only baseline by $1.15$ to $3.92$ mIoU points. On an AgML agricultural split, SegRAG uses an average of $28.8$ annotated references per class, raises mean IoU from $25.27$ to $59.24$, and recovers classes where text-only grounding scores zero.

\section{Research Objectives}
\label{sec:objectives}

This study has four research objectives.

\begin{enumerate}
  \item \textbf{RO1:} design a mechanism that augments a frozen promptable segmentation model with external visual evidence without updating its weights.
  \item \textbf{RO2:} determine whether annotated reference images can be converted into a compact and reliable class memory without fine-tuning DINOv3 or SAM\,3.
  \item \textbf{RO3:} test whether retrieval-derived spatial prompts improve segmentation beyond text-only prompting on standard open-vocabulary benchmarks and on domain-shifted agricultural imagery.
  \item \textbf{RO4:} identify which components of the pipeline, including bank filtering, spatial peak selection, text prompting, and descriptor backbone, are responsible for the observed gains.
\end{enumerate}

The main contribution is a retrieval-based prompt formation pipeline rather than a new segmentation backbone. SegRAG differs from prior exemplar- and retrieval-based methods in three ways. First, it filters the reference memory before inference by checking whether candidate descriptors retrieve same-class foreground on the other selected support images. Second, it converts retrieved evidence into localized point prompts before SAM\,3 decoding, rather than classifying masks after they have already been proposed. Third, it combines class text and retrieval-derived points in one model state, so semantic intent and spatial evidence constrain the same mask prediction.

\section{Related Work and Research Gap}
\label{sec:related}

\subsection{Open-vocabulary segmentation and the grounding gap}

Semantic segmentation has progressed from fixed-label dense predictors, including FCNs, U-Net, DeepLab, and Mask2Former~\cite{long2015fcn,ronneberger2015unet,chen2018deeplabv3plus,cheng2022mask2former}, to systems that can respond to categories expressed at inference time. Vision--language pretraining enabled this transition by aligning images with natural-language concepts~\cite{radford2021clip}, and open-vocabulary methods adapted those representations to dense prediction through prompt learning, mask classification, and feature refinement~\cite{li2022lseg,ghiasi2022openseg,liang2023ovseg,yu2023fcclip,xu2023odise}. CLIP refinements improve localization without task-specific retraining~\cite{hajimiri2024naclip,lan2024proxyclip,zhang2025corrclip}, but their spatial predictions remain tied to what the pretrained representation associates with the requested text.

Promptable foundation models separate object delineation from a fixed output vocabulary. SAM and SAM\,2 accept geometric prompts~\cite{kirillov2023sam,ravi2025sam2}, while SAM\,3 also supports concept prompts for open-vocabulary mask generation~\cite{carion2025sam3}. This broadens the set of categories that can be requested, but it does not ensure that a class name is grounded in the correct image region when its appearance differs from the model's pretrained concept. The unresolved problem is therefore not only vocabulary coverage. It is how to supply reliable target-domain spatial evidence without updating the foundation model.

\subsection{Visual exemplars and external memory}

Retrieval-augmented systems use external evidence when fixed model parameters do not contain, or do not expose, the information required for a query~\cite{lewis2020rag,khandelwal2020knnlm}. In vision, retrieval has conditioned generation~\cite{blattmann2022rdm,chen2023reimagen} and supported open-vocabulary recognition and segmentation~\cite{shin2022reco,gui2024knnclip}. Few-shot segmentation methods such as Matcher, IPSeg, GF-SAM, and VRP-SAM demonstrate that annotated supports can provide class-specific visual evidence~\cite{liu2024matcher,tang2024ipseg,zhang2024gfsam,sun2024vrpsam}. Their supports, however, are primarily episode-level inputs. They do not address the separate information-management problem of maintaining a persistent class memory while deciding which local descriptors remain trustworthy across references.

Other approaches change the source or placement of external evidence. FOSSIL and FreeDA construct prototypes from synthetic imagery, avoiding manual support annotation at the cost of possible mismatch with the target acquisition domain~\cite{barsellotti2024fossil,barsellotti2024freeda}. No Time to Train! stores reference descriptors and uses them to label masks after automatic SAM proposal generation~\cite{espinosa2025notimetotrain}. This is effective when the required region is already proposed, but retrieval cannot guide the formation of a proposal that was absent from the unguided mask set.

Taken together, prior work provides semantic grounding, exemplar evidence, and retrieval-based transfer, but leaves two linked issues underexplored: curating reliable evidence across annotated references and injecting it before mask formation. SegRAG addresses this gap with a persistent class-level descriptor memory and ICCD filtering, then uses TSG to retrieve spatial locations and convert them into point prompts before SAM\,3 decoding, allowing the stored evidence to guide mask formation together with the class text. It is thus a pre-decoding, non-parametric grounding mechanism rather than a new segmentation backbone, keeping domain knowledge inspectable and updateable outside model weights.

\section{Methodology}
\label{sec:methodology}

\begin{figure}[!t]
    \centering
    \includegraphics[width=\textwidth]{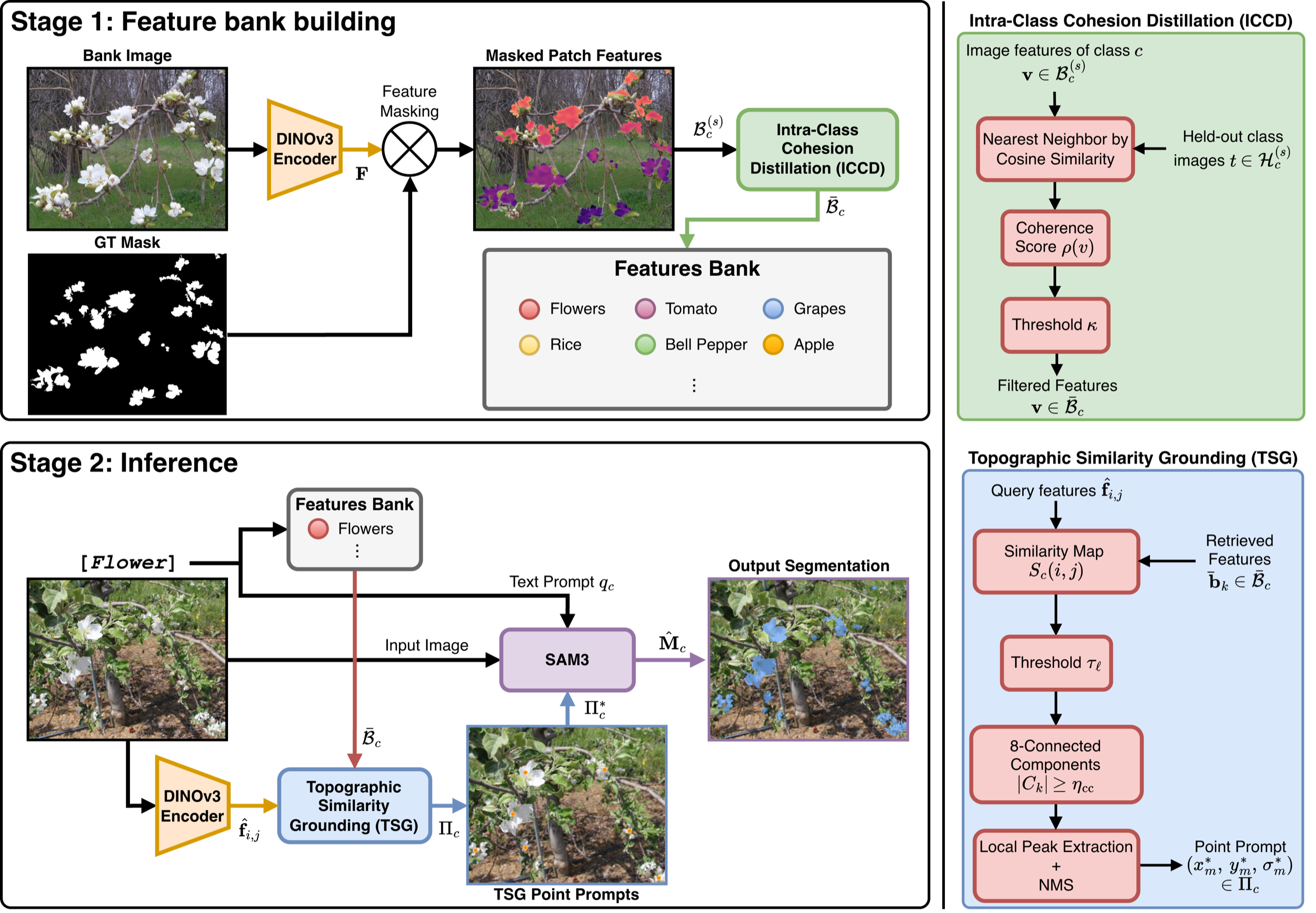}
    \caption{Overview of the proposed retrieval-supported segmentation workflow.
    Stage~I (top) builds a class-keyed descriptor bank from annotated
    references and, when multiple references are available, filters it with
    Intra-Class Cohesion Distillation (ICCD). Stage~II (bottom) matches query
    features against this bank using Topographic Similarity Grounding (TSG) to
    produce point prompts, which are combined with the class text and passed
    to SAM\,3 in a single grounding call to produce the predicted mask
    $\hat{\mathbf{M}}_c$. Full notation and thresholds are defined in
    Sections~\ref{sec:stage1} and~\ref{sec:stage2}.}

    \label{fig:pipeline}
\end{figure}

SegRAG separates persistent class evidence from query-specific localization. A vetted repository of DINOv3~\cite{simeoni2025dinov3} descriptors supplies class-aware point cues tied to image locations, and SAM\,3~\cite{carion2025sam3} uses those cues together with the class text to produce the mask. Figure~\ref{fig:pipeline} depicts the two-stage workflow. In the offline stage, labelled support imagery is converted into a descriptor store keyed by class name. During inference, the requested category selects the corresponding class-level bank. All retained descriptors in that bank are then matched against the query image to identify likely spatial evidence for the class, after which the textual class prompt and the retrieved point evidence are provided together to SAM\,3 for mask decoding.

\subsection{Stage I: Feature Bank Construction}
\label{sec:stage1}

Subsequent query matching requires a bank that is selective rather than merely large. Stage~I therefore produces a concise category-specific descriptor set by first encoding labelled references densely and assigning their foreground patch vectors to class-name entries, then, when multiple references are available, applying cross-image screening to remove vectors whose retrieval behaviour is not stable for their category.

\subsubsection{Dense Feature Extraction}
\label{sec:feature_extraction}

\paragraph{Image preprocessing.}
Cosine comparisons across reference and query imagery are meaningful only when all DINOv3 inputs share the same numerical convention. For this reason, a fixed preprocessing map $\mathcal{T}$ is applied before DINOv3 encoding, whether the image is used for bank construction or for query-time inference. This transform is specific to DINOv3, and SAM\,3 instead receives the original RGB image through its own image processor. For an image $\mathcal{I}$, the operation resizes the input to $1536 \times 1536$ pixels with bicubic interpolation and then performs ImageNet-channel normalisation:
\begin{equation}
\tilde{\mathcal{I}} = \mathcal{T}(\mathcal{I}) = \frac{\operatorname{BicubicResize}(\mathcal{I},\,1536\times1536) - \boldsymbol{\mu}_{\text{IN}}}{\boldsymbol{\sigma}_{\text{IN}}},
\label{eq:preprocess}
\end{equation}
where $\boldsymbol{\mu}_{\text{IN}} = (0.485, 0.456, 0.406)$ and $\boldsymbol{\sigma}_{\text{IN}} = (0.229, 0.224, 0.225)$ give the ImageNet channel means and standard deviations. Applying exactly the same $\mathcal{T}$ in both stages keeps the descriptor distribution aligned. Otherwise, small preprocessing mismatches would alter feature magnitudes and directions, weakening the cosine retrieval operations on which the pipeline relies.

\paragraph{Backbone feature extraction.}
The preprocessed image $\tilde{\mathcal{I}}$ is encoded by a DINOv3
ViT-L/16 backbone $\phi$ whose parameters remain fixed. We discard the
class token and the four register tokens, retain only the patch tokens
from the final transformer block after the model's internal layer
normalisation, and reshape them to a two-dimensional patch lattice. With
input side length $1536$ and patch size $P = 16$, this lattice has
$H_g = W_g = \lfloor 1536 / P \rfloor = 96$ locations and descriptor
dimension $D = 1024$:
\begin{equation}
\mathbf{F} = \phi(\tilde{\mathcal{I}}) \in \mathbb{R}^{H_g \times W_g \times D}.
\label{eq:dinov3}
\end{equation}
At lattice coordinate $(i,j)$, the descriptor is written as $\mathbf{f}_{i,j} \in \mathbb{R}^D$, and its unit-length version is $\hat{\mathbf{f}}_{i,j} = \mathbf{f}_{i,j} / \|\mathbf{f}_{i,j}\|_2$. All descriptors are stored and compared in unquantized \texttt{float32}, ensuring that every module operates in the same numerical representation.

\paragraph{Mask-to-patch alignment and raw bank construction.}
Patch inclusion is determined by foreground occupancy rather than by a single pixel decision. Let $\mathcal{S}_c$ denote the set of reference images carrying an annotation for class $c$. For each image $s \in \mathcal{S}_c$, the annotation is decoded as a binary mask $\mathbf{M}_c^{(s)} \in \{0,1\}^{H_s \times W_s}$. To match the DINOv3 patch lattice, this mask is first resized to $1536 \times 1536$ using nearest-neighbour interpolation and then averaged over non-overlapping $P \times P$ blocks with stride $P$:
\begin{equation}
\tilde{M}_{c,i,j}^{(s)} = \frac{1}{P^2} \sum_{h=iP}^{(i+1)P-1} \sum_{w=jP}^{(j+1)P-1} M_{c,h,w}^{(s)},
\label{eq:avgpool}
\end{equation}
for $i=0,\ldots,H_g-1$ and $j=0,\ldots,W_g-1$, where $M_{c,h,w}^{(s)}$ is the value at pixel location $(h,w)$ in the resized version of $\mathbf{M}_c^{(s)}$. A patch becomes a foreground patch for class $c$ only if the object covers more than the high-occupancy cutoff $\tau_b$:
\begin{equation}
\mathcal{F}_b\!\left(\mathbf{M}_c^{(s)}\right) = \bigl\{(i,j) : \tilde{M}_{c,i,j}^{(s)} > \tau_b\bigr\}.
\label{eq:foreground_bank}
\end{equation}
Using a large $\tau_b$ deliberately suppresses boundary patches, so the raw bank receives descriptors sampled from inside the labelled object rather than from mixed foreground--background areas. The class-specific subset contributed by image $s$ is
\begin{equation}
\mathcal{B}_c^{(s)} = \left\{ \hat{\mathbf{f}}_{i,j}^{(s)} : (i,j) \in \mathcal{F}_b\!\left(\mathbf{M}_c^{(s)}\right) \right\},
\label{eq:per_image_bank}
\end{equation}
and the unfiltered bank for class $c$ is formed by combining the subsets over all $|\mathcal{S}_c|$ references:
\begin{equation}
\mathcal{B}_c = \biguplus_{s \in \mathcal{S}_c} \mathcal{B}_c^{(s)}.
\label{eq:raw_bank}
\end{equation}
Here $\biguplus$ denotes a source-indexed union: each stored descriptor retains the identifier of the image from which it was extracted. Accordingly, $s(\mathbf{v})$ denotes the source image associated with a descriptor $\mathbf{v} \in \mathcal{B}_c$. Since every stored vector has unit norm, later dot products are cosine similarities without any additional conversion.

\subsubsection{Intra-Class Cohesion Distillation (ICCD)}
\label{sec:iccd}

\begin{figure}[!t]
    \centering
    \includegraphics[width=0.8\linewidth]{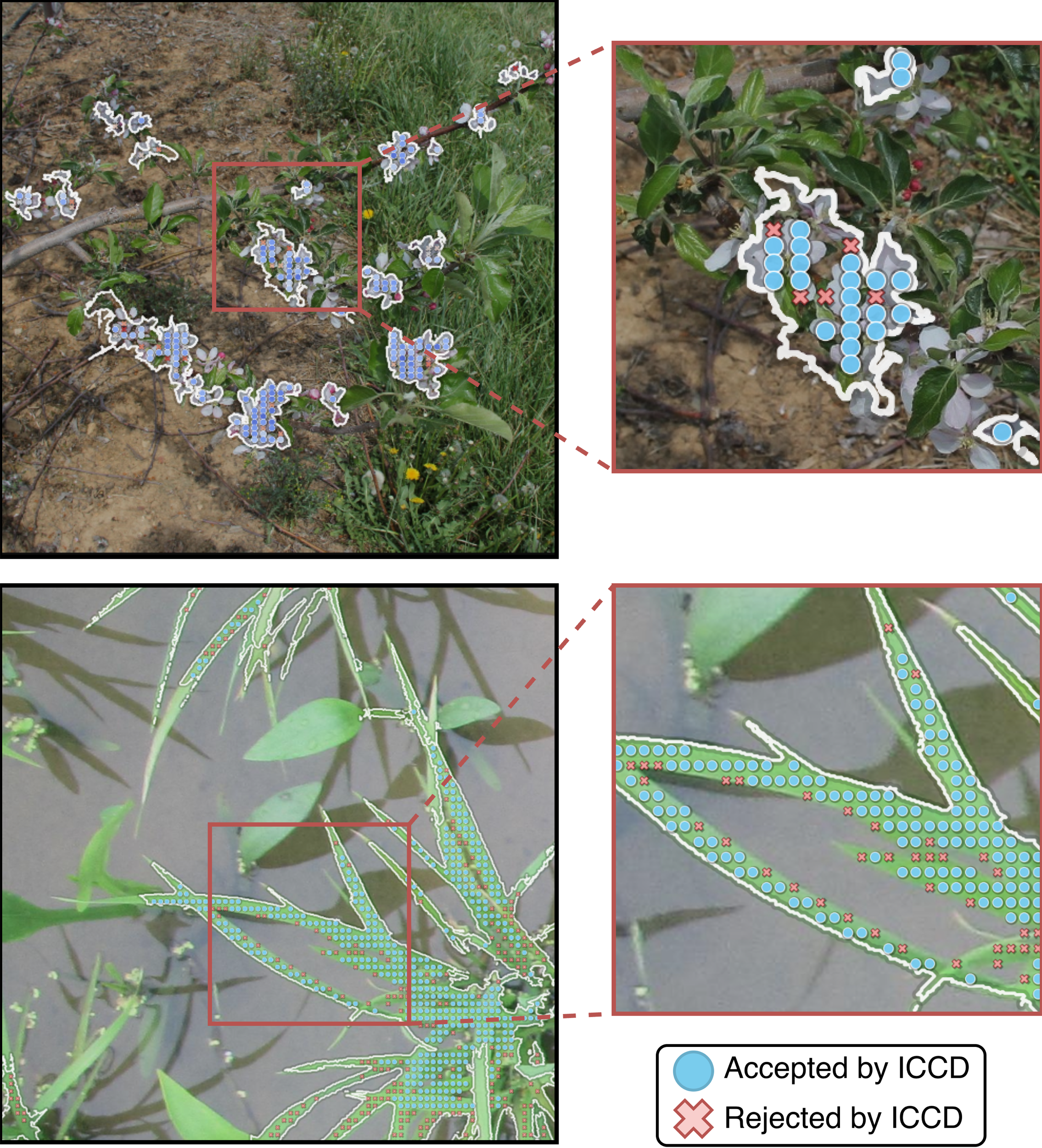}
    \caption{ICCD filtering results for two example categories
    (top: small clustered flowers, bottom: rice leaves). Blue circles denote
    patch descriptors retained in $\bar{\mathcal{B}}_c$, whereas red crosses
    denote descriptors removed by the filtering rule. The rejected descriptors
    are concentrated around three error sources: \emph{boundary patches},
    \emph{fuzzy-mask patches}, and \emph{appearance-ambiguous patches}.}
    \label{fig:iccd}
\end{figure}

The high-occupancy mask gate in Eq.~\eqref{eq:foreground_bank} improves descriptor purity, but it does not guarantee category selectivity. Figure~\ref{fig:iccd} shows three common sources of unreliable bank entries: \emph{boundary patches} that mix object and background content, \emph{fuzzy-mask patches} caused by annotations that include semantically mixed areas, such as visible background between leaves, and \emph{appearance-ambiguous patches} whose descriptors are not sufficiently different from nearby background. ICCD is designed to reduce these cases through a retrieval-reliability score assigned to each candidate vector. The score measures whether that vector retrieves foreground locations for the same class in the other selected support images. Candidates showing unstable cross-image behaviour, imprecise mask support, or poor separation from background are removed, whereas vectors with stable within-class retrieval behaviour remain in $\bar{\mathcal{B}}_c$.

\paragraph{Target mask construction.}
Self-comparison would overstate descriptor quality, so a vector sampled from image $s$ is evaluated only on the complementary set $\mathcal{H}_c^{(s)} = \mathcal{S}_c \setminus \{s\}$. For a target image $t \in \mathcal{H}_c^{(s)}$, the foreground map used for scoring is produced with a more permissive occupancy cutoff $\tau_t < \tau_b$, giving retrieved patches a tolerant but still class-aware evaluation mask:
\begin{equation}
\mathcal{F}_t\!\left(\mathbf{M}_c^{(t)}\right) = \bigl\{p : \tilde{M}_{c,p}^{(t)} > \tau_t\bigr\},
\label{eq:foreground_target}
\end{equation}
where $p$ indexes the flattened set of $N_p = H_g W_g$ patch locations. The binary indicator $\mathbf{m}^{(t)} \in \{0,1\}^{N_p}$ is defined so that $m_p^{(t)} = 1$ exactly when $p \in \mathcal{F}_t(\mathbf{M}_c^{(t)})$. During ICCD, patches are referenced by the flat index $p$. The coordinate form $(i,j)$ used during bank creation in Section~\ref{sec:feature_extraction} is only an equivalent two-dimensional notation for the same $N_p$ positions.

\paragraph{Nearest-neighbour assignment and similarity gating.}
A target image contributes evidence for a source vector only if the closest target patch is similar enough to be meaningful. Consider $\mathbf{v} \in \mathcal{B}_c^{(s)}$ from Eq.~\eqref{eq:per_image_bank} and a target image $t \in \mathcal{H}_c^{(s)}$. Let $\hat{\mathbf{G}}^{(t)} \in \mathbb{R}^{N_p \times D}$ be the flattened target patch matrix obtained with the same $\phi$ and the same normalisation as the source descriptors. Its row descriptors are denoted $\hat{\mathbf{g}}_p^{(t)}$ rather than $\hat{\mathbf{f}}_p^{(t)}$ only to distinguish target-image descriptors from source-side bank or query descriptors. All are $\ell_2$-normalised outputs of $\phi$. The strongest correspondence in $t$ is
\begin{equation}
p^*(\mathbf{v}, t) = \arg\max_{p \in \{1,\ldots,N_p\}} \mathbf{v} \cdot \hat{\mathbf{g}}_p^{(t)},
\label{eq:nn_assign}
\end{equation}
and the dot product is a cosine similarity because both vectors have unit norm. Low-similarity correspondences are not treated as reliable evidence, so a minimum cosine gate $\xi \geq 0$ defines the subset of target images that are \emph{valid} for scoring $\mathbf{v}$:
\begin{equation}
\mathcal{V}(\mathbf{v}) = \bigl\{ t \in \mathcal{H}_c^{(s)} : \mathbf{v} \cdot \hat{\mathbf{g}}_{p^*(\mathbf{v},t)}^{(t)} \geq \xi \bigr\}.
\label{eq:valid_targets}
\end{equation}
Targets below this gate are omitted because such weak correspondences cannot distinguish genuine class retrieval from incidental background similarity. With $\mathbb{1}[\cdot]$ denoting an indicator, valid targets are counted according to whether their best-matching patch falls inside or outside the target foreground:
\begin{align}
\text{good}(\mathbf{v}) &= \sum_{t \in \mathcal{V}(\mathbf{v})} \mathbb{1}\!\left[m_{p^*(\mathbf{v},t)}^{(t)} = 1\right], \label{eq:good} \\[4pt]
\text{bad}(\mathbf{v})  &= \sum_{t \in \mathcal{V}(\mathbf{v})} \mathbb{1}\!\left[m_{p^*(\mathbf{v},t)}^{(t)} = 0\right]. \label{eq:bad}
\end{align}
Each valid image supplies exactly one nearest-neighbour assignment, hence $\text{good}(\mathbf{v}) + \text{bad}(\mathbf{v}) = |\mathcal{V}(\mathbf{v})|$. Unlike $\mathcal{H}_c^{(s)}$, the size of $\mathcal{V}(\mathbf{v})$ depends on the descriptor itself: vectors that match many targets with high confidence receive more scoring evidence than vectors whose best correspondences are weak.

\paragraph{Intra-class coherence score.}
The score for a descriptor is only trusted when enough valid target images participate. This requirement is enforced through the required-match count $\eta_{\min}$, and the coherence value is then the fraction of valid assignments that land on the foreground of class $c$:
\begin{equation}
\rho(\mathbf{v}) = \begin{cases} \dfrac{\text{good}(\mathbf{v})}{|\mathcal{V}(\mathbf{v})|} & \text{if } |\mathcal{V}(\mathbf{v})| \geq \eta_{\min}, \\[8pt] -1 & \text{otherwise.} \end{cases}
\label{eq:coherence}
\end{equation}
Descriptors with fewer than $\eta_{\min}$ valid correspondences receive $\rho(\mathbf{v})=-1$. They are therefore barred from the refined bank, not because they are known to be incorrect, but because the available cross-image evidence is too sparse for a dependable estimate.

\paragraph{Adaptive coherence threshold.}
A single global cutoff would impose the same standard on classes with very different visual diversity and difficulty. Instead, the eligibility threshold is estimated separately for each class. Before this class-specific estimation, a fixed preliminary floor $\lambda = 0.60$ removes descriptors whose retrieval precision is not even minimally reliable. This floor is a coarse, class-independent pre-filter and is not a substitute for the class-specific step that follows. The selected support set $\mathcal{S}_c$ is used both to construct $\mathcal{B}_c$ and to provide cross-image scoring evidence. The set $\Psi_c$ collects candidate vectors from this bank whose coherence score clears the preliminary floor:
\begin{equation}
\Psi_c = \bigl\{ \mathbf{v} \in \mathcal{B}_c : \rho(\mathbf{v}) \geq \lambda \bigr\}, \qquad \lambda = 0.60.
\label{eq:scored_set}
\end{equation}
The set of scalar scores is $R_c = \{\rho(\mathbf{v}) : \mathbf{v} \in \Psi_c\} \subseteq [\lambda,1]$, and $Q_{3/4}(R_c)$ denotes its 75th percentile. This third-quartile summary reflects the stronger part of the class-specific score distribution while avoiding dependence on isolated extreme values. The class-specific cutoff is then
\begin{equation}
\kappa_c = \operatorname{clip}\!\left(0.90 \cdot Q_{3/4}(R_c),\; \kappa_{\mathrm{lo}},\; \kappa_{\mathrm{hi}}\right),
\label{eq:adaptive_threshold}
\end{equation}
where $\operatorname{clip}(x, a, b) = \max(a, \min(b, x))$, with floor $\kappa_{\mathrm{lo}} = 0.65$ and ceiling $\kappa_{\mathrm{hi}} = 0.82$. The factor $0.90$ relaxes the third-quartile score slightly, allowing high-quality but not necessarily peak-ranked descriptors to survive. The lower bound $\kappa_{\mathrm{lo}}$ prevents visually weak classes from admitting many unreliable vectors when their scores are broadly low, while the upper bound $\kappa_{\mathrm{hi}}$ prevents overly aggressive pruning for classes whose descriptors are already highly separable.

\paragraph{Top-$K$ filtering and bank refinement.}
After scoring, the refined descriptor set is constrained by both the adaptive cutoff and the capacity limit $K$. The following path applies when $|\mathcal{S}_c| \geq 2$, so that each source image has at least one comparison image. The $|\mathcal{S}_c| = 1$ case is handled by the contingency path below. Support images are processed deterministically in sorted image-identifier order. For every descriptor, the scoring targets are the remaining $|\mathcal{S}_c|-1$ selected supports after excluding its source image. A vector is eligible when its coherence score reaches the class-specific threshold from Eq.~\eqref{eq:adaptive_threshold}:
\begin{equation}
\mathcal{E}_c = \bigl\{ \mathbf{v} \in \mathcal{B}_c : \rho(\mathbf{v}) \geq \kappa_c \bigr\},
\label{eq:eligible}
\end{equation}
where $\rho(\mathbf{v})$ is evaluated against $\mathcal{H}_c^{(s(\mathbf{v}))}$. If at most $K$ eligible descriptors remain, all are stored. Otherwise, the descriptors are ranked by $\rho(\mathbf{v})$. Let $\operatorname{TopK}_K(\mathcal{E}_c;\rho)$ denote the subset containing the $K$ descriptors in $\mathcal{E}_c$ with the largest coherence scores. The refined bank is
\begin{equation}
\bar{\mathcal{B}}_c = \begin{cases}
\mathcal{E}_c, & \text{if } |\mathcal{E}_c| \leq K, \\[4pt]
\operatorname{TopK}_K(\mathcal{E}_c;\rho), & \text{otherwise.}
\end{cases}
\label{eq:filtered_bank}
\end{equation}
yielding $|\bar{\mathcal{B}}_c| = \min(|\mathcal{E}_c|, K)$. The refined bank preserves the class-name key of the raw bank and is written to disk so that inference can retrieve it by category name.

\paragraph{Single-reference contingency.}
ICCD is a cross-image reliability estimate and is only defined when at least two reference images are available for a class. If $|\mathcal{S}_c|=1$, the complementary set $\mathcal{H}_c^{(s)}$ is empty, so $\rho(\mathbf{v})$ is not computed, and neither the preliminary floor $\lambda$ nor the class-specific cutoff $\kappa_c$ is applied. The one-reference setting therefore uses a simpler contingency path: descriptors selected by the high-occupancy foreground gate in Eq.~\eqref{eq:foreground_bank} are retained as $\bar{\mathcal{B}}_c$, subject only to the same capacity cap $K$ and deterministic ordering used by the implementation. Because this occupancy gate already enforces a high foreground-purity requirement ($\tau_b=0.90$), the retained descriptors remain strongly interior even without additional coherence scoring. This path preserves the information supplied by the annotated mask while omitting cross-image scoring, which is unavailable from a single reference.

This contingency does not assign a single-image reliability score and does not estimate cross-image coherence. It is therefore not treated as an ICCD estimate.

\subsection{Stage II: Inference}
\label{sec:stage2}

Stage~II turns a query image $\mathcal{I}$ and a class name $c$ into a grounded mask by combining retrieved category evidence with SAM\,3 decoding. The class key first selects the entire corresponding class-level bank $\bar{\mathcal{B}}_c$. Query features are then matched to those descriptors to form a dense response map, TSG converts that map into point prompts, and SAM\,3 receives both the text prompt and the point set within one shared grounding invocation.

\subsubsection{Topographic Similarity Grounding (TSG)}
\label{sec:tsg}

\begin{figure}[!htbp]
  \centering
  \includegraphics[width=\linewidth]{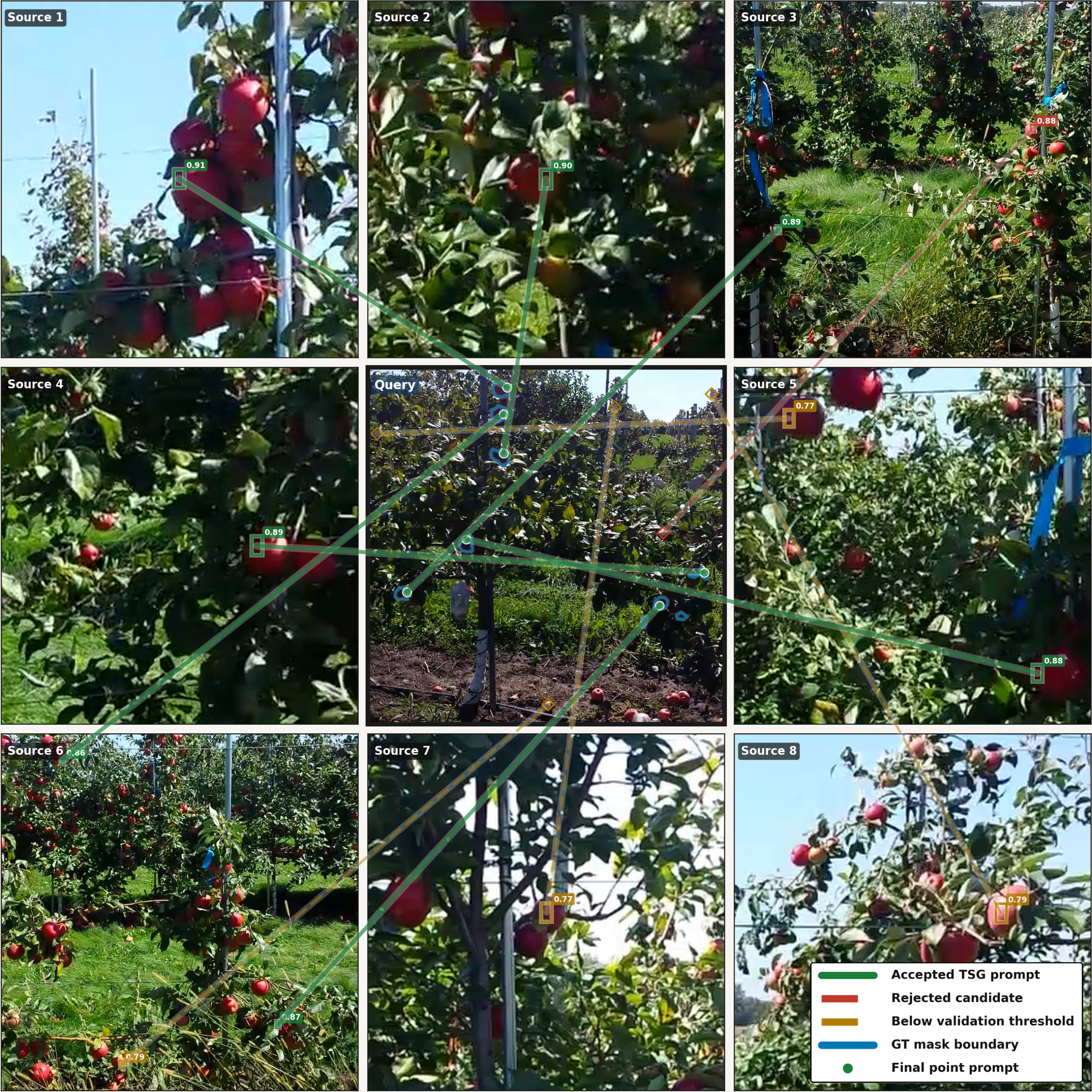}
  \caption{
    Illustration of Topographic Similarity Grounding (TSG) for the class \texttt{apple}. Eight representative reference images contributing descriptors to $\bar{\mathcal{B}}_c$ are shown around the query image at the centre. This number is illustrative, since $|\mathcal{S}_c|$ is class-dependent in practice. Each coloured line joins a bank patch, annotated with its cosine score $S_c(i,j)$, to the nearest-neighbour location found in the query feature grid. \textbf{Green lines} indicate matches whose peak value reaches $\tau_v$ and whose connected component satisfies the minimum-size gate $\eta_{\text{cc}}$. These matches supply the foreground point prompts passed to SAM\,3. \textbf{Orange lines} denote proposals that pass the loose cutoff $\tau_\ell$ but do not reach $\tau_v$, while \textbf{red} markers correspond to patches removed by connected-component filtering in Eq.~\eqref{eq:cc_filter}. Blue contours show the ground-truth mask boundaries. Raw similarity responses, contiguous components, and validated peaks form the successive TSG filtering cascade described in Section~\ref{sec:tsg}.}
  \label{fig:tsg}
\end{figure}

The purpose of Topographic Similarity Grounding (TSG) is to convert retrieved descriptors into image-location prompt coordinates for the query. Rather than accepting every patch whose score crosses a threshold, TSG reads the spatial arrangement of cosine responses: contiguous high-response zones are found first, and representative maxima are selected from within them. This topographic view supports images with multiple instances because $n$ separated objects of class $c$ tend to create $n$ separate response modes, each of which can yield its own point prompt. Figure~\ref{fig:tsg} presents this process for \texttt{apple}: Eq.~\eqref{eq:simmap} defines the scores $S_c(i,j)$, match lines connect reference-bank patches to their nearest query locations, and the colour progression from raw responses in red, to coherent candidates in orange, to final validated prompts in green circles corresponds to the three steps below.

\paragraph{Similarity map computation.}
For the query image, the same $\mathcal{T}$ and $\phi$ described in Section~\ref{sec:feature_extraction} produce $\mathbf{F} \in \mathbb{R}^{H_g \times W_g \times D}$. Query descriptors are written without an image superscript to distinguish them from the source descriptors $\hat{\mathbf{f}}_{i,j}^{(s)}$ and scoring-target descriptors $\hat{\mathbf{g}}_p^{(t)}$ used in Stage~I. The retrieved class bank is written as $\bar{\mathcal{B}}_c = \{\hat{\mathbf{b}}_1, \ldots, \hat{\mathbf{b}}_{|\bar{\mathcal{B}}_c|}\}$. Its entries are already unit vectors because each was inserted as a normalised patch descriptor $\hat{\mathbf{f}}_{i,j}^{(s)}$ in Eq.~\eqref{eq:per_image_bank}. The hats on $\hat{\mathbf{b}}_k$ emphasize that fact. At each query patch coordinate $(i,j)$, the largest cosine match over the retrieved prototypes is
\begin{equation}
S_c(i,j) = \max_{k \in \{1,\ldots,|\bar{\mathcal{B}}_c|\}} \hat{\mathbf{f}}_{i,j} \cdot \hat{\mathbf{b}}_k,
\label{eq:simmap}
\end{equation}
which gives the response map $\mathbf{S}_c \in \mathbb{R}^{H_g \times W_g}$. Large responses indicate regions whose local features resemble the class prototypes, so the map encodes both likely class location and retrieval confidence. Because this spatial prior is obtained entirely from retrieved reference descriptors without task-specific spatial supervision, $\mathbf{S}_c$ acts as the \emph{topographic similarity landscape} for class $c$.

\paragraph{Connected-component candidate filtering.}
A patchwise threshold alone would admit isolated false activations as prompts. TSG instead filters the response field spatially, using the shape of $\mathbf{S}_c$ before any point prompt is emitted.

Applying the \emph{loose} cutoff $\tau_\ell$ produces the permissive binary proposal map
\begin{equation}
\mathcal{L}_c = \bigl\{(i,j) : S_c(i,j) \geq \tau_\ell\bigr\}.
\label{eq:loose_mask}
\end{equation}
An 8-connected labelling procedure divides $\mathcal{L}_c$ into non-overlapping regions $\{C_1, \ldots, C_N\}$ such that $\mathcal{L}_c = \bigcup_{k=1}^{N} C_k$ and $C_k \cap C_{k'} = \emptyset$ for all $k \neq k'$. Regions smaller than $\eta_{\text{cc}}$ patches are removed:
\begin{equation}
\mathcal{C}_c = \bigl\{C_k : k \in \{1,\ldots,N\},\ |C_k| \geq \eta_{\text{cc}}\bigr\}.
\label{eq:cc_filter}
\end{equation}
Thus, each $C_k \in \mathcal{C}_c$ is a contiguous proposal zone that is likely to contain at least one instance of class $c$, whereas $\eta_{\text{cc}}$ blocks isolated single-patch detections that a direct thresholding rule would otherwise pass to the prompt generator.

\paragraph{Per-component peak extraction.}
Within a retained component, prompts should be placed at the strongest regional evidence rather than across the whole component. For each $C_k$, the response map is masked outside the component:
\begin{equation}
\tilde{S}_c^{(k)}(i,j) = \begin{cases} S_c(i,j) & \text{if } (i,j) \in C_k, \\ -\infty & \text{otherwise,} \end{cases}
\label{eq:masked_simmap}
\end{equation}
and maxima are detected using a window of size $(2\delta+1)\times(2\delta+1)$, with $\delta$ representing the required separation between neighbouring peaks in patch units. A patch $(i,j)$ qualifies as a local maximum when
\begin{equation}
\tilde{S}_c^{(k)}(i,j) = \max_{(i',j') \in \mathcal{N}_\delta(i,j)} \tilde{S}_c^{(k)}(i',j'),
\label{eq:peak_cond}
\end{equation}
where $\mathcal{N}_\delta(i,j) = \{(i',j') : \max(|i'-i|, |j'-j|) \leq \delta\}$ is the Chebyshev-metric neighbourhood of radius $\delta$. The overall maximum of a non-empty component necessarily satisfies Eq.~\eqref{eq:peak_cond}, since no neighbour has a larger value, so every such component contributes at least one maximum before suppression.
For each component, candidate maxima are sorted by decreasing $S_c(i,j)$ and processed by greedy Euclidean non-maximum suppression (NMS) using the same distance parameter $\delta$. A candidate is kept precisely when no earlier accepted maximum lies within distance $\delta$. Since the first candidate is the highest-scoring one and is accepted before comparisons with lower-ranked candidates, a non-empty candidate list always yields at least one retained peak. As a defensive fallback, if the list entering NMS is empty---a condition that cannot arise under normal operation but is guarded against for robustness---the highest-scoring patch in $C_k$ is used as the prompt location.
All accepted peaks from all components in $\mathcal{C}_c$ are pooled and sorted by decreasing $S_c$. If a downstream prompt allowance $B_{\max}$ is provided at prediction time, only the top-$B_{\max}$ entries are used. Otherwise, the collection is left uncapped. After this optional cap, let $N_c^{\mathrm{p}}$ denote the number of retained peaks, enumerated as $(i_m^*,j_m^*)$ for $m=1,\ldots,N_c^{\mathrm{p}}$ in descending response order. Each peak is mapped to the centre of its corresponding image patch:
\begin{equation}
x_m^* = \Bigl(j_m^* + \tfrac{1}{2}\Bigr)\frac{W}{W_g}, \qquad
y_m^* = \Bigl(i_m^* + \tfrac{1}{2}\Bigr)\frac{H}{H_g},
\quad m=1,\ldots,N_c^{\mathrm{p}},
\label{eq:pixel_coords}
\end{equation}
where $W$ and $H$ are the original image width and height. The score of the $m$-th prompt is $\sigma_m^* = S_c(i_m^*, j_m^*) \in [-1,1]$, since cosine similarity is bounded by $[-1,1]$. The ranked prompt set is therefore
\begin{equation}
\Pi_c = \bigl\{(x_m^*,\, y_m^*,\, \sigma_m^*)\bigr\}_{m=1}^{N_c^{\mathrm{p}}},
\label{eq:prompt_set}
\end{equation}
with entries arranged by descending $\sigma^*$. By construction, every element of $\Pi_c$ corresponds to a regional mode in the topographic similarity landscape. Thus, when $n$ spatially separated objects of class $c$ appear, the response map usually forms $n$ distinct modes above the loose cutoff, allowing TSG to produce separate prompts for separate instances.

\FloatBarrier
\subsubsection{Simultaneous Multimodal Prompting}
\label{sec:multimodal_prompting}

SAM\,3 receives the linguistic cue and the TSG-derived point cues in the same decoder state. They are not processed as independent inference calls. Instead, both modalities are inserted before decoding, so semantic intent and spatial evidence influence the mask during one forward invocation.

\paragraph{Prompt construction.}
Let $\gamma_c$ be the canonical class name, such as \texttt{dog} or \texttt{sofa}. Underscores in $\gamma_c$ are replaced with spaces, and the article is selected from the first letter of the resulting string. Writing juxtaposition as string concatenation with a separating space, the text prompt is
\begin{equation}
q_c = a(\gamma_c)\;\gamma_c,
\label{eq:text_prompt}
\end{equation}
where $a(\gamma_c) \in \{\text{``a''},\, \text{``an''}\}$ equals ``an'' when $\gamma_c$ begins with a vowel and ``a'' otherwise. From $\Pi_c$ in Eq.~\eqref{eq:prompt_set}, only prompts whose scores satisfy the validation cutoff $\tau_v \in (0,1)$ are retained:
\begin{equation}
\Pi_c^* = \bigl\{(x_m^*,\, y_m^*,\, \sigma_m^*) \in \Pi_c : \sigma_m^* \geq \tau_v\bigr\}.
\label{eq:validated_prompts}
\end{equation}
Because $\tau_v$ lies in $(0,1)$, each accepted point has positive cosine evidence. Moreover, each peak in $\Pi_c$ came from a component $C_k \subseteq \mathcal{L}_c$, and membership in $\mathcal{L}_c$ requires $S_c(i,j) \geq \tau_\ell$. Hence every peak already obeys $\sigma_m^* \geq \tau_\ell$. It is therefore appropriate to choose $\tau_v \geq \tau_\ell$, making $\tau_v$ a true confidence filter rather than a redundant pass-through. In all reported SegRAG experiments, including both the standard benchmarks and the agricultural ablations (Section~\ref{sec:ablation}), the values are set equal, $\tau_v = \tau_\ell = 0.80$ (Section~\ref{sec:setup}), so every peak entering $\Pi_c$ also passes validation. The two parameters nonetheless remain conceptually distinct and need not be tied together: a practitioner may raise $\tau_v$ above $\tau_\ell$ to tighten point-level confidence without altering how candidate regions are delineated, without needing to re-run region formation.
The two thresholds operate at different stages. The loose threshold $\tau_\ell$ shapes the topographic response field into contiguous candidate zones, whereas $\tau_v$ is imposed after peak extraction as an absolute confidence requirement on individual prompt scores. Accepted coordinates are then converted to the unit image domain required by SAM\,3's processor:
\begin{equation}
\tilde{x}_m = x_m^* / W, \qquad \tilde{y}_m = y_m^* / H,
\label{eq:norm_coords}
\end{equation}
and placed in $\mathbf{P}_c \in \mathbb{R}^{|\Pi_c^*| \times 1 \times 2}$, with $[\mathbf{P}_c]_{m,1,:} = (\tilde{x}_m,\, \tilde{y}_m)$ for $m = 1,\ldots,|\Pi_c^*|$. A label tensor $\mathbf{1}_c \in \mathbb{R}^{|\Pi_c^*| \times 1}$ accompanies the coordinates, with $[\mathbf{1}_c]_{m,1}=1$ for every retained prompt to indicate foreground.

\paragraph{Joint state construction.}
The decoder state is assembled before any mask prediction is requested. Let $\Omega_0$ denote a newly initialized empty SAM\,3 inference state. Through SAM\,3's state-carrying interface, image features, language embeddings, and point constraints are placed into a class-specific state $\Omega_c$. The image is encoded once:
\begin{equation}
\mathbf{E} = \textsc{SAM3-Encode}(\mathcal{I}),
\label{eq:image_enc}
\end{equation}
producing $\mathbf{E}$ for later decoding. The class text $q_c$ is then processed by SAM\,3's language encoder:
\begin{equation}
\mathbf{T}_c = \textsc{SAM3-TextEncode}(q_c),
\label{eq:text_enc}
\end{equation}
which yields $\mathbf{T}_c$. After the image and text embeddings are available, the TSG point coordinates and foreground labels are inserted into the same state:
\begin{equation}
\Omega_c = \textsc{Update}\!\left(\Omega_0,\, \mathbf{E},\, \mathbf{T}_c,\, \mathbf{P}_c,\, \mathbf{1}_c\right).
\label{eq:state_update}
\end{equation}

\paragraph{Single-pass joint grounding.}
Mask generation then requires one decoder invocation after $\Omega_c$ has been populated with both semantic and geometric inputs:
\begin{equation}
\hat{\mathbf{M}}_c = \textsc{SAM3-Ground}(\Omega_c).
\label{eq:joint_grounding}
\end{equation}
Within SAM\,3, the text prompt conditions the upstream fusion encoder so that image features are semantically aligned before localization, while the point prompts provide explicit spatial constraints to the mask decoder. Since both signals are already present in $\Omega_c$ at decode time, they jointly determine the output mask rather than operating as a text-first step followed by point-based correction. This distinguishes the formulation from a sequential design in which textual grounding and point refinement are executed as separate stages.

\paragraph{Graceful degradation.}
A prediction should still be produced if no TSG point satisfies Eq.~\eqref{eq:validated_prompts} or if the joint call is incompatible with the internal state. In either case, the system falls back to text-only decoding:
\begin{equation}
\hat{\mathbf{M}}_c = \textsc{SAM3-Ground}\!\left(\textsc{Update}(\Omega_0,\, \mathbf{E},\, \mathbf{T}_c)\right),
\label{eq:fallback}
\end{equation}
This fallback guarantees a mask output and preserves the recoverability of the text-prompt baseline whenever geometric prompting cannot be used.

\section{Datasets and Experimental Protocol}
\label{sec:datasets_protocol}

\subsection{Dataset suite}
\label{sec:datasets}

The experiments use five dataset groups. Four are standard open-vocabulary segmentation benchmarks (see Table~\ref{tab:datasets}). The fifth, AgML \cite{joshi2023agml}, is used as a domain-transfer stress test because agricultural imagery contains sensor, viewpoint, growth-stage, and crop-variety shifts that are poorly represented by generic web imagery~\cite{lei2024review}.

\begin{table}[!t]
\centering
\caption{Datasets used in the evaluation. The table separates the benchmark role from the way each dataset is used by SegRAG.}
\label{tab:datasets}
\setlength{\tabcolsep}{5pt}
\renewcommand{\arraystretch}{1.18}
\resizebox{\textwidth}{!}{%
\begin{tabular}{p{2.4cm}p{3.2cm}p{4.2cm}p{4.6cm}}
\toprule
\textbf{Dataset} & \textbf{Evaluation role} & \textbf{Scale and labels} & \textbf{Use in this study} \\
\midrule
ADE20K-150~\cite{zhou2017ade20k} & Scene parsing across varied indoor and outdoor images & $150$ semantic categories, commonly evaluated on $2{,}000$ validation images after training on $20{,}210$ images & One and five annotated references per class are used to construct SegRAG banks. mIoU is reported on the validation split. \\
Cityscapes ~\cite{cordts2016cityscapes} & High-resolution urban scene segmentation & Fine annotations cover $2{,}975$ training, $500$ validation, and $1{,}525$ test images, with $19$ common evaluation classes & One and five reference images per class are used for the banks. mIoU is reported on the validation split. \\
PC-59~\cite{mottaghi2014pc59} & Object and context segmentation in crowded natural scenes & PASCAL Context provides dense scene annotations. The PC-59 protocol evaluates $59$ frequent categories & One and five reference images per class are used for the banks. mIoU is reported under the PC-59 protocol. \\
LVIS~\cite{gupta2019lvis} & Long-tailed large-vocabulary instance segmentation & More than $1{,}200$ categories, about $164{,}000$ images, and more than $2$ million masks & One and five reference images per class are used for bank construction. mIoU is reported for the compared category set. \\
AgML~\cite{joshi2023agml} & Agricultural domain transfer & Eleven foreground classes are evaluated: apple, bean leaf, bell pepper, carrot, cauliflower, flower, grape, rice, sugarbeet weed, tomato, and weed. The class-level evaluation manifests contain $2{,}183$ image entries & The main up-to-30-shot result uses $317$ class-level reference image entries, or $28.82$ per class on average. Ten classes use $30$ references each, and sugarbeet weed uses all $17$ available training images. The support-selection analysis uses one and five references per class. \\
\bottomrule
\end{tabular}%
}
\end{table}

The public dataset releases and their official splits are used; no new images were collected for this study. Reference candidates are drawn from the corresponding training split and must contain an annotation for the target class. Evaluation follows the validation or test protocol stated in Table~\ref{tab:datasets}. Ignore regions and LVIS non-exhaustive annotations are handled according to the official evaluation rules. For SegRAG, the preprocessing described in Section~\ref{sec:feature_extraction} is applied during both bank construction and query matching. The experiments retain exact support and query manifests, which are checked for the required support counts, class and query coverage, and complete prediction accounting.
The AgML evaluation is class stratified. For each of the eleven foreground classes, annotated references are used only to construct the feature bank. No model weights are updated with AgML images. For the main up-to-30-shot result, support images are selected from the training split as the first $30$ sorted image identifiers containing the target class, where available. The same selected $N$ images construct and score the bank, and each source is scored against only the other $N-1$ supports. Ten classes use $N=30$, while sugarbeet weed uses all $N=17$ available supports. Evaluation uses every test image containing the class, and no query subsampling is applied.

An audit of the AgML support and query manifests found no identical file reuse, but identified five of the $2{,}183$ query--class records, corresponding to four canonical source frames, that shared source identifiers with support records. Excluding these records changed mIoU by only $0.02$ points under the audited evaluation, indicating negligible sensitivity to the detected shared provenance.

\subsection{Baselines, metrics, and implementation}
\label{sec:setup}

Mean Intersection over Union (mIoU) is the main metric. The evaluation includes text-only open-vocabulary methods, synthetic-prototype methods, prompt-driven multimodal large language model (MLLM) segmentation, retrieval-based systems, exemplar-based SAM pipelines, and SegRAG. Five reference-based methods were evaluated locally where their released implementations permit one- or five-shot evaluation. The five-shot comparisons with GF-SAM, FSSDINO, IPSeg, and Matcher on ADE20K-150, Cityscapes, and PC-59 use the same logical support and query memberships, evaluated classes, inclusion and ignore rules, prediction accounting, and class-aggregated semantic-mIoU implementation as SegRAG. Each method retains its released backbone, decoder, preprocessing, and prompt construction, so this is a matched data and evaluation protocol rather than an architecture-controlled comparison. Other locally evaluated cells and scores taken directly from the original papers are presented as contextual and are distinguished in Table~\ref{tab:main_comparison}. SAM\,3 text-only provides the direct controlled comparison that keeps the decoder family fixed and isolates the addition of retrieval-derived spatial prompts.
Sa2VA-4B is evaluated through its official interface on the fixed PC-59 and AgML query manifests. Multiple returned masks are combined by union, and a missing mask is scored as an empty prediction. Support-selection sensitivity is evaluated on PC-59 and AgML over five paired seeds ($17$, $29$, $43$, $71$, and $101$). Supports are sampled uniformly without replacement from the training split after sorting by class and image identifiers. The query set is held fixed, and each one-shot support is nested in the corresponding five-shot set.
For the reported DINOv3 SegRAG runs, support and query images passed to DINOv3 are resized to $1536 \times 1536$ using bicubic interpolation and ImageNet normalisation. SAM\,3 receives the original RGB image through its own processor. Binary annotation masks are resized using nearest-neighbour interpolation. Raw-bank descriptors require foreground occupancy $\tau_b=0.90$, whereas ICCD target masks use the more permissive occupancy threshold $\tau_t=0.50$. Multi-reference ICCD uses the preliminary coherence floor $\lambda=0.60$, cosine-validity threshold $\xi=0.0$, and required-match count $\eta_{\min}=3$. For an $N$-shot condition, the same selected $N$ supports construct the bank and provide cross-image evidence, with each source excluded from its own $N-1$ scoring targets. The one-reference path bypasses these coherence-scoring and adaptive-filtering operations. The descriptor-bank capacity is $K=10{,}000$ per class. TSG uses $\tau_\ell=\tau_v=0.80$, minimum connected-component size $\eta_{\mathrm{cc}}=4$, and NMS distance $\delta=10$ patch units unless an ablation explicitly varies one of these values. DINOv3 ViT-L/16 and SAM\,3 remain frozen in all SegRAG experiments. Here, training-free means that no model weights are updated and no gradient-based training is performed, annotated reference masks are still required to construct the class memory.

\section{Results}
\label{sec:results}

\subsection{Standard benchmark comparison}
\label{sec:results_standard}

Table~\ref{tab:main_comparison} reports the standard benchmark results. SegRAG is evaluated with only one or five annotated references per class on these datasets. Double-daggered five-shot cells identify the four baselines included in the matched-data comparison on ADE20K-150, Cityscapes, and PC-59. Within this scope, SegRAG is higher than all four methods on each dataset. Relative to the direct SAM\,3 text-only baseline, five-shot SegRAG improves by $2.51$ mIoU on ADE20K-150, $2.47$ on Cityscapes, $1.15$ on PC-59, and $3.92$ on LVIS. These gains are obtained without weight updates and with only five annotated references per class. The remaining local and literature-reported cells provide context but do not expand the matched-data claim.

\begin{table*}[t]
\centering
\caption{%
  Comparison with open-vocabulary and training-free semantic segmentation methods on
  four standard benchmarks.
  All scores are mIoU~(\%) on the standard validation split. \nd{} denotes not
  reported by the original paper and not evaluated in this work.
  Few-shot and $N$-shot bank scores are reported as \emph{1-shot\,/\,5-shot}.
  \textbf{Setting}: \emph{text}~=~text prompt only.
  \emph{1-shot} or \emph{1\,/\,5-shot}~=~1 or 5 annotated reference images per
  class passed at inference (exemplar) or used to build the bank ($N$-shot bank).
  \emph{bank\,(full)}~=~offline feature bank built from the entire training set.
  \textbf{SegRAG} augments SAM\,3 with a DINOv3 feature bank and TSG-derived point prompts. The one-shot bank uses occupancy-selected raw descriptors, whereas multi-reference banks are filtered by ICCD.
  SAM\,3 is therefore both the base model and the direct controlled baseline.
  A double dagger marks a five-shot value included in the matched-data comparison on ADE20K-150, Cityscapes, or PC-59.
  $\Delta$~columns show the gain of \textbf{SegRAG}~(5-shot) over SAM\,3 text-only prompting.
}
\label{tab:main_comparison}
\resizebox{\textwidth}{!}{%
\setlength{\tabcolsep}{4pt}
\begin{tabular}{l l l l c c c c c c c c}
\toprule
\multirow{2}{*}{Method} &
\multirow{2}{*}{Venue} &
\multirow{2}{*}{Backbone} &
\multirow{2}{*}{Setting} &
\multicolumn{2}{c}{ADE20K-150} &
\multicolumn{2}{c}{Cityscapes} &
\multicolumn{2}{c}{PC-59} &
\multicolumn{2}{c}{LVIS} \\
\cmidrule(lr){5-6}\cmidrule(lr){7-8}\cmidrule(lr){9-10}\cmidrule(lr){11-12}
& & & &
mIoU & $\Delta$ &
mIoU & $\Delta$ &
mIoU & $\Delta$ &
mIoU & $\Delta$ \\
\midrule

\multicolumn{12}{l}{\textit{Attention-modified CLIP}} \\[2pt]

NACLIP~\cite{hajimiri2024naclip}
  & WACV 2025 & CLIP & text
  & $17.4^{\dagger}$ & & $35.5^{\dagger}$ & & $35.2^{\dagger}$ & & \nd & \\

ProxyCLIP~\cite{lan2024proxyclip}
  & ECCV 2024 & CLIP\,+\,DINOv2 & text
  & $20.2^{\dagger}$ & & $38.1^{\dagger}$ & & $39.1^{\dagger}$ & & \nd & \\

CorrCLIP~\cite{zhang2025corrclip}
  & ICCV 2025 & CLIP\,+\,DINOv2\,+\,SAM\,2 & text
  & $26.9^{\dagger}$ & & $49.4^{\dagger}$ & & $48.8^{\dagger}$ & & \nd & \\

\midrule

\multicolumn{12}{l}{\textit{Text-only open-vocabulary}} \\[2pt]

SAM\,3~\cite{carion2025sam3}
  & Meta AI 2025 & PE\,+\,SAM\,3 & text
  & \underline{52.26} & & \underline{64.78} & & \underline{65.62} & & \underline{54.92} & \\

Grounded SAM~\cite{ren2024groundedsam}
  & arXiv 2024 & Grounding DINO\,+\,SAM & text
  & 48.76 & & 47.41 & & 62.38 & & 47.33 & \\

\midrule

\multicolumn{12}{l}{\textit{Full or synthetic bank (training-free)}} \\[2pt]

kNN-CLIP~\cite{gui2024knnclip}
  & arXiv 2024 & DINOv2\,+\,CLIP & full bank
  & $41.3^{\dagger}$ & & \nd & & $62.8^{\dagger}$ & & \nd & \\

FOSSIL~\cite{barsellotti2024fossil}
  & WACV 2024 & SD\,+\,DINOv2 & synthetic
  & $18.8^{\dagger}$ & & $23.2^{\dagger}$ & & $35.8^{\dagger}$ & & \nd & \\

FreeDA~(ViT-L)~\cite{barsellotti2024freeda}
  & CVPR 2024 & SD\,+\,DINOv2 & synthetic
  & $23.2^{\dagger}$ & & $36.7^{\dagger}$ & & $43.5^{\dagger}$ & & \nd & \\

\midrule

\multicolumn{12}{l}{\textit{Few-shot exemplar (training-free, support passed at inference)}} \\[2pt]

Matcher~\cite{liu2024matcher}
  & ICLR 2024
  & DINOv2\,+\,SAM
  & 1\,/\,5-shot
  & \nd\,/$35.15^{\ddagger}$ & & \nd\,/$23.94^{\ddagger}$ & & 48.06\,/$47.26^{\ddagger}$ & & 33.0\,/\,$40.0^{\dagger}$ & \\

GF-SAM~\cite{zhang2024gfsam}
  & NeurIPS 2024
  & DINOv2\,+\,SAM\,2
  & 1\,/\,5-shot
  & 43.66\,/$50.65^{\ddagger}$ & & 35.17\,/$40.04^{\ddagger}$ & & 53.29\,/$62.21^{\ddagger}$ & & 35.2\,/\,$44.2^{\dagger}$ & \\

FSSDINO~\cite{zakir2026fssdino}
  & arXiv 2026
  & DINOv3
  & 1\,/\,5-shot
  & 33.53\,/$38.79^{\ddagger}$ & & 17.38\,/$18.50^{\ddagger}$ & & 39.31\,/$48.33^{\ddagger}$ & & \nd\,/\,32.39 & \\

IPSeg~\cite{tang2024ipseg}
  & IJCV 2025
  & DINOv2\,+\,SD\,+\,SAM
  & 1\,/\,5-shot
  & 31.45\,/$32.76^{\ddagger}$ & & 20.17\,/$20.58^{\ddagger}$ & & 42.77\,/$44.83^{\ddagger}$ & & \nd\,/\,\nd & \\

\midrule

\multicolumn{12}{l}{\textit{$N$-shot bank (training-free, offline bank of $N$ references per class)}} \\[2pt]

No Time to Train!~\cite{espinosa2025notimetotrain}
  & arXiv 2025
  & DINOv2\,+\,SAM\,2
  & 1\,/\,5-shot
  & 35.19\,/\,42.45 & & 34.49\,/\,44.40 & & 43.83\,/\,56.42 & & \nd\,/\,\nd & \\

\rowcolor{rowours}
\textbf{SegRAG (ours)}
  & ---
  & DINOv3\,+\,SAM\,3
  & 1\,/\,5-shot
  & 53.52\,/\,$\mathbf{54.77}^{\ddagger}$ & \pos{2.51}
  & 66.35\,/\,$\mathbf{67.25}^{\ddagger}$ & \pos{2.47}
  & 65.91\,/\,$\mathbf{66.77}^{\ddagger}$ & \pos{1.15}
  & 56.71\,/\,\textbf{58.84} & \pos{3.92} \\

\bottomrule
\end{tabular}%
}

\smallskip
\noindent\footnotesize
${}^{\dagger}$~Score taken directly from the original paper. ${}^{\ddagger}$~Five-shot score included in the matched-data comparison. Unmarked few-shot scores were evaluated locally but are presented as contextual unless otherwise stated.

\end{table*}

The closest bank-based comparison is No Time to Train!~\cite{espinosa2025notimetotrain}. At five shots, SegRAG is higher by $12.32$ mIoU on ADE20K-150, $22.85$ on Cityscapes, and $10.35$ on PC-59. No Time to Train! first uses SAM\,2 to generate candidate masks, then matches DINOv2 features from those masks against reference prototypes to classify and rank the proposals. Its reference memory therefore selects among existing candidates rather than directly guiding where masks are proposed. SegRAG instead retrieves query locations before SAM\,3 decoding and turns them into point prompts, allowing the stored features to contribute directly to locating instances, including those missed by text-only grounding. The comparison is not fully architecture controlled because the methods use different SAM generations. However, the DINOv2--DINOv3 ablation in Section~\ref{sec:ablation_backbone} changes SegRAG's AgML mIoU by only $0.42$ points while keeping its pipeline and SAM\,3 fixed, indicating that DINOv3 alone does not explain the domain-transfer result.
Within the matched five-shot comparison, the nearest margins are $4.12$ mIoU on ADE20K-150, $27.21$ on Cityscapes, and $4.56$ on PC-59, each against GF-SAM. These results show that SegRAG remains effective with a small annotated support set under matched logical data and evaluation accounting. The competing methods retain different backbones, decoders, preprocessing, and prompt construction, so their comparisons establish competitiveness rather than isolate a single architectural cause. SAM\,3 text-only provides the direct controlled comparison for the effect of adding retrieved spatial evidence.
\subsection{Agricultural domain transfer}
\label{sec:agri}

Table~\ref{tab:agri_comparison} reports the agricultural domain-transfer stress test. SAM\,3 text-only prompting obtains $25.27$ mean IoU across eleven AgML classes. The up-to-30-shot SegRAG condition, which uses $28.82$ references per class on average, reaches $59.24$, a gain of $33.97$ points. Four classes score zero under text-only prompting and become positive under SegRAG. Cauliflower rises from $0.00$ to $95.36$ IoU and sugarbeet weed rises from $0.00$ to $80.22$ IoU.

\begin{table}[!t]
\centering
\caption{Per-class IoU for the AgML up-to-30-shot evaluation. The same selected $N$ supports build and score each class bank, with self-comparison excluded. Ten classes use 30 supports and sugarbeet weed uses 17. A zero value for SAM\,3 indicates complete text-only grounding failure under the tested imaging conditions.}
\label{tab:agri_comparison}
\setlength{\tabcolsep}{8pt}
\renewcommand{\arraystretch}{1.2}
\begin{tabular}{l c c c}
\toprule
\textbf{Class} & \textbf{SAM\,3} & \textbf{SegRAG (ours)} & \textbf{Gain} \\
\midrule
apple           & 37.27 & \textbf{37.65} & \textcolor{posgreen}{$+0.38$}  \\
bean leaf       & 4.81  & \textbf{63.91} & \textcolor{posgreen}{$+59.10$} \\
bell pepper     & 74.21 & \textbf{81.16} & \textcolor{posgreen}{$+6.95$}  \\
carrot          & \cellcolor{red!20}0.00 & \textbf{19.90} & \textcolor{posgreen}{$+19.90$} \\
cauliflower     & \cellcolor{red!20}0.00 & \textbf{95.36} & \textcolor{posgreen}{$+95.36$} \\
flower          & 31.36 & \textbf{40.68} & \textcolor{posgreen}{$+9.32$}  \\
grape           & 54.93 & \textbf{71.21} & \textcolor{posgreen}{$+16.28$} \\
rice            & \cellcolor{red!20}0.00 & \textbf{39.93} & \textcolor{posgreen}{$+39.93$} \\
sugarbeet weed  & \cellcolor{red!20}0.00 & \textbf{80.22} & \textcolor{posgreen}{$+80.22$} \\
tomato          & 67.12 & \textbf{67.80} & \textcolor{posgreen}{$+0.68$}  \\
weed            & 8.26  & \textbf{53.78} & \textcolor{posgreen}{$+45.52$} \\
\midrule
\rowcolor{rowours}\textbf{Mean} & \textbf{25.27} & \textbf{59.24} & \textcolor{posgreen}{$+33.97$} \\
\bottomrule
\end{tabular}
\end{table}

The per-class pattern is informative. Apple and tomato improve only slightly because their common appearance is already represented well enough for a text prompt to work. Bean leaf, weed, cauliflower, sugarbeet weed, rice, and carrot require stronger spatial evidence. In these cases, the reference bank contains local appearances that the text prompt cannot express. TSG then supplies coordinates for the decoder, which converts retrieved appearance evidence into a mask.

\subsection{Comparison with Sa2VA-4B}
\label{sec:mllm_comparison}

To place SegRAG against a contemporary prompt-driven MLLM, we evaluate Sa2VA-4B~\cite{sa2va} on the fixed PC-59 and AgML query sets. This comparison is intended as a stress test rather than a like-for-like architectural comparison. SegRAG is a segmentation-specific pipeline built from frozen vision models and does not include an LLM, whereas Sa2VA is a substantially heavier general-purpose multimodal language model with segmentation capability. The computational profile in Section~\ref{sec:computational_cost} gives $1.16$B loaded parameters for SegRAG and $3.94$B for Sa2VA, so SegRAG uses about $29\%$ of Sa2VA's parameter count. Sa2VA also requires no annotated support images, while SegRAG uses target-domain references. The comparison therefore reflects different deployment choices and information access.

\begin{table}[H]
\centering
\caption{Stress-test comparison with Sa2VA-4B. All scores are mIoU (\%). SegRAG values for PC-59 and AgML are means and sample standard deviations over five independently sampled five-shot support sets. The final column reports the up-to-30-shot AgML setting, with 30 references for ten classes and 17 for sugarbeet weed.}
\label{tab:sa2va}
\setlength{\tabcolsep}{7pt}
\renewcommand{\arraystretch}{1.12}
\begin{tabular}{lccc}
\toprule
\textbf{Model} & \textbf{PC-59} & \textbf{AgML 5-shot} & \textbf{AgML main} \\
\midrule
Sa2VA-4B & $\mathbf{68.49}$ & $50.14$ & $50.14$ \\
\rowcolor{blue!10}
SegRAG (ours) & $66.63\pm0.17$ & $\mathbf{58.51\pm1.30}$ & $\mathbf{59.24}$ \\
\midrule
Difference & \textcolor{red}{$-1.86$} & \textcolor{posgreen}{$+8.37$} & \textcolor{posgreen}{$+9.10$} \\
\bottomrule
\end{tabular}%
\end{table}

Sa2VA is stronger on PC-59 despite receiving no annotated support images. The ranking reverses on AgML, where SegRAG is higher by $8.37$ points under repeated five-shot sampling and by $9.10$ points in the main setting. This pattern indicates that SegRAG has better domain transfer performance, despite using only $29\%$ of Sa2VA's parameter count. The result does not establish universal superiority across all conditions, but it shows that a smaller, segmentation-specific retrieval pipeline can outperform a considerably larger general-purpose MLLM on domain transfer when annotated target-domain evidence is available.

\subsection{Computational complexity}
\label{sec:computational_cost}

Table~\ref{tab:complexity} reports a controlled computational comparison on the same $62$-image, $118$ image-class PC-59 workload. All methods are evaluated on one RTX 3090 with batch size one. Online timing excludes model loading, metric scoring, and disk output and is reported as mean and sample standard deviation over three full passes. FLOPs are dynamically counted over registered PyTorch operations, with one multiply-accumulate counted as two FLOPs. Peak memory is the maximum in-process allocated VRAM. SegRAG is measured in uncached online mode.

\begin{table}[H]
\centering
\caption{Controlled online complexity and offline memory-preparation cost for SAM\,3 text-only~\cite{carion2025sam3}, No Time to Train! (NTTT)~\cite{espinosa2025notimetotrain}, Sa2VA-4B~\cite{sa2va}, and SegRAG. Param. denotes loaded parameters, GFLOPs is the counted computation per image, Time is online latency in milliseconds per image, VRAM is peak allocated memory in gigabytes, and Prep. is offline memory-preparation time in seconds. A dash indicates that no external reference memory is constructed. The NTTT offline value includes memory filling and post-processing.}
\label{tab:complexity}
\setlength{\tabcolsep}{4pt}
\renewcommand{\arraystretch}{1.12}
\begin{tabular}{lccccc}
\toprule
\textbf{Model} & \textbf{Param.} & \textbf{GFLOPs} & \textbf{Time} & \textbf{VRAM} & \textbf{Prep.} \\
\midrule
SAM\,3 text-only & $860.06$M & $11{,}941$ & $289.2\pm0.5$ & $5.49$ & -- \\
NTTT 5-shot & $528.80$M & $6{,}809$ & $1{,}267.9\pm3.9$ & $12.20$ & $99.53$ \\
Sa2VA-4B & $3.94$B & $54{,}240$ & $1{,}889.3\pm7.2$ & $9.89$ & -- \\
\rowcolor{rowours}\textbf{SegRAG 5-shot} & $1.16$B & $20{,}702$ & $1{,}349.3\pm1.4$ & $5.67$ & $501.69$ \\
\bottomrule
\end{tabular}
\end{table}

SAM\,3 text-only is the fastest condition because it performs no query-feature matching or retrieval. SegRAG is $6.4\%$ slower than No Time to Train! in the uncached online profile, but it uses $53.6\%$ less peak VRAM. Relative to Sa2VA-4B, SegRAG uses $70.5\%$ fewer loaded parameters, $61.8\%$ fewer counted FLOPs, $42.7\%$ less peak VRAM, and is $28.6\%$ faster. These measurements reinforce that the Sa2VA comparison is a stress test across models of different scope rather than a like-for-like comparison. For SegRAG, approximately $1{,}159.9$ ms/image is spent on DINOv3 query-feature extraction and TSG, followed by $189.5$ ms/image for simultaneous SAM\,3 text-and-point decoding.

The one-time SegRAG five-shot feature-bank build takes $501.69$ s, peaks at $4.18$ GiB host RAM and $3.21$ GiB allocated VRAM, and is excluded from online latency. The resulting five-shot bank occupies $0.568$ GB. Together with the SAM\,3 and DINOv3 checkpoints, the deployment footprint is $5.231$ GB, compared with $15.159$ GB for the Sa2VA checkpoint. Thus, retrieval introduces measurable offline and online cost, but the resulting deployment remains substantially smaller than the evaluated MLLM. The appropriate conclusion is an accuracy--efficiency trade-off rather than universal speed superiority.

\subsection{Qualitative analysis}
\label{sec:qualitative}

\begin{figure}[!t]
    \centering
    \includegraphics[width=\textwidth]{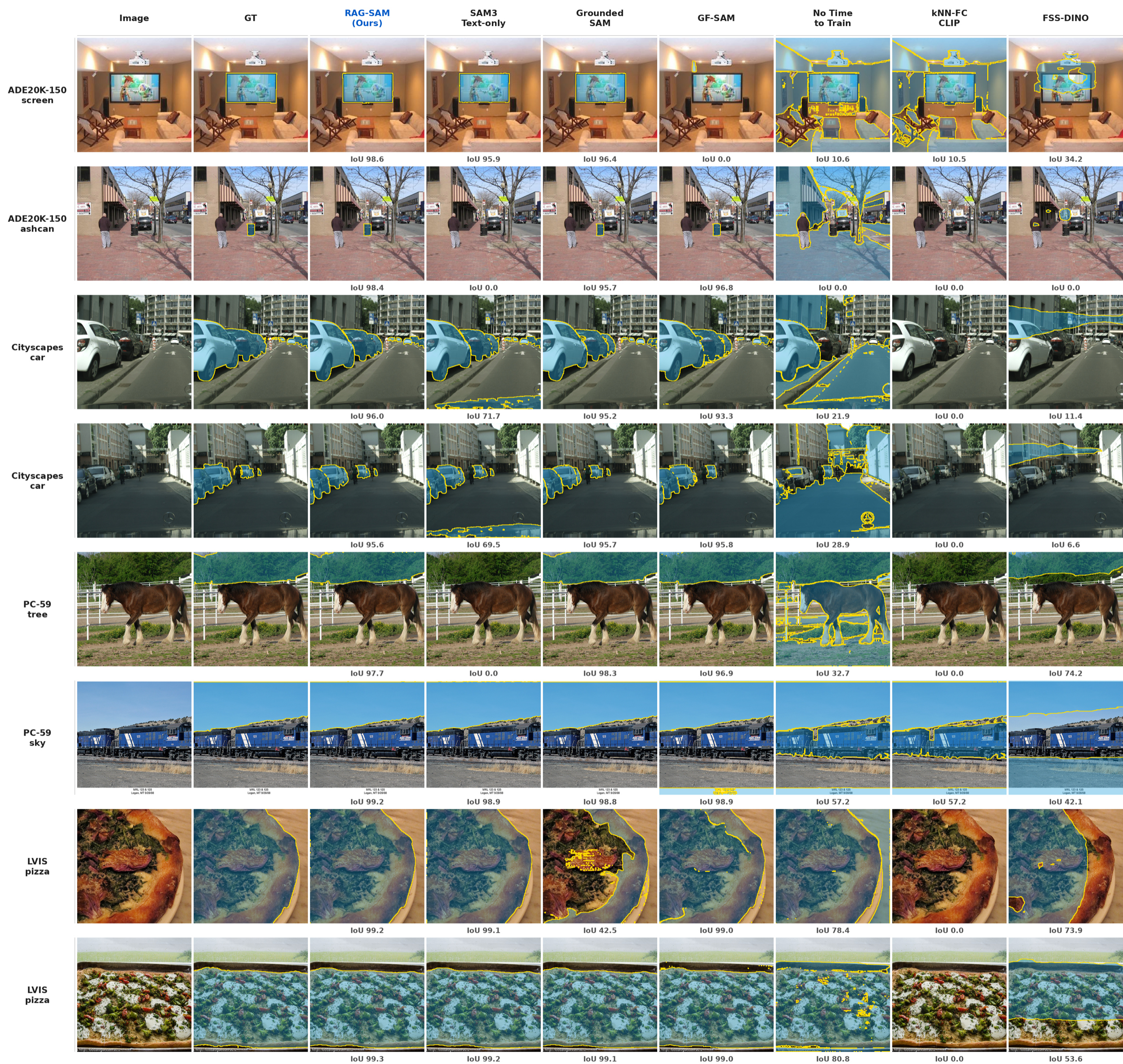}
    \caption{Qualitative comparison on the four standard benchmarks. The examples show that SegRAG changes masks most when the target is small, visually subtle, or split across instances.}
    \label{fig:all_models_comparison}
\end{figure}

Figure~\ref{fig:all_models_comparison} shows two modes. In easy cases, such as a large sky region or a salient pizza instance, SegRAG remains close to the already strong SAM\,3 output. In harder cases, such as small ashcan regions or multiple cars in Cityscapes, retrieved points give the decoder spatial anchors that text-only grounding lacks. This behavior is consistent with the quantitative results, where the largest standard benchmark gain appears on LVIS, the dataset with the strongest long-tail pressure.

\begin{figure}[!t]
    \centering
    \includegraphics[width=\textwidth]{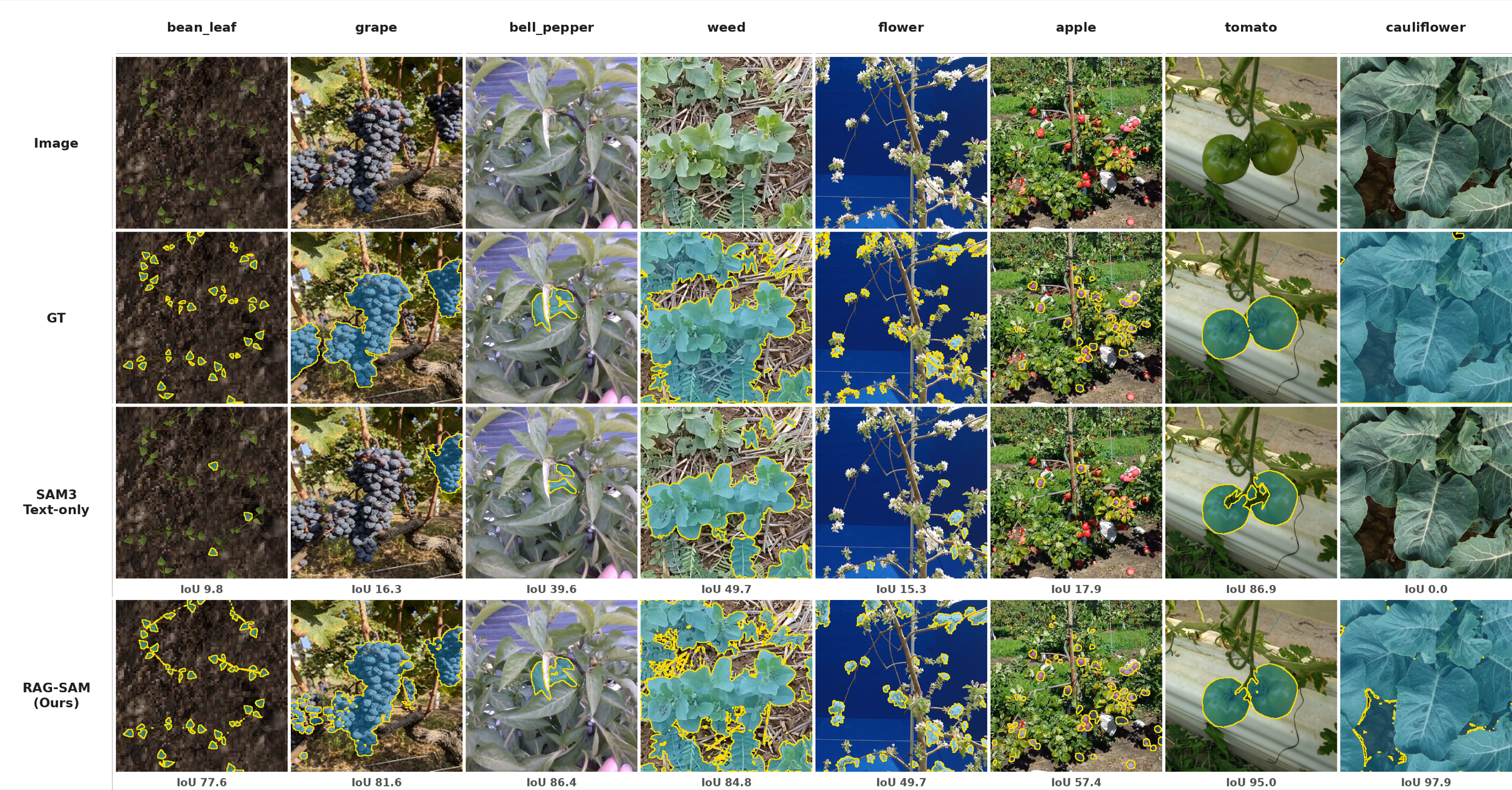}
    \caption{Qualitative comparison on AgML agricultural examples. SAM\,3 often misses targets whose field appearance differs from a common web depiction. SegRAG uses annotated references to localize those appearances.}
    \label{fig:agri_comparison}
\end{figure}

Figure~\ref{fig:agri_comparison} illustrates the agricultural failure mode. Young cauliflower plants, scattered bean leaves, occluded bell peppers, and mixed weed regions do not provide a reliable text-only anchor. SegRAG improves these cases because the retrieved descriptors come from the target imaging domain and because TSG selects separate prompt locations for separated instances.

\section{Ablation Analysis}
\label{sec:ablation}

Unless stated otherwise, the ablations use the eleven AgML classes, $\tau_\ell=\tau_v=0.80$, $\eta_{\mathrm{cc}}=4$, and $\delta=10$ patch units. The primary component experiment uses the controlled seed-17 five-shot protocol, while each secondary sensitivity table states its own reference setting. This evaluation is intentionally difficult because text-only grounding fails on several classes, making the contribution of each component visible.

\subsection{Controlled Component and Prompt Ablation}
\label{sec:ablation_components}

The primary ablation uses one frozen protocol throughout: the seed-17 five-shot AgML supports, $2{,}183$ fixed query--class pairs over eleven classes, the same DINOv3 and SAM\,3 checkpoints, and the same prediction and metric implementations. Support-dependent rows use identical support images and masks. Table~\ref{tab:ablation_components} varies bank curation, point selection, and decoder prompt modality within this protocol.

\begin{table*}[!t]
\centering
\caption{Controlled component and prompt-modality ablation on AgML. All rows use the seed-17 five-shot protocol, $2{,}183$ query--class pairs over eleven classes, fixed DINOv3 and SAM\,3 checkpoints, and the same mask-generation and metric implementations. A check mark enables a component or prompt modality, and a cross disables it. In point-based rows, disabling ICCD uses the raw foreground bank, while disabling TSG supplies every dense query patch passing the fixed $0.80$ similarity threshold without connected-component filtering, peak extraction, or spatial NMS. All values are percentages. Bold marks the best value and underlining marks the second-best value in each metric column.}
\label{tab:ablation_components}
\setlength{\tabcolsep}{5pt}
\renewcommand{\arraystretch}{1.12}
\resizebox{\textwidth}{!}{%
\begin{tabular}{c|cccc|ccccc}
\toprule
\textbf{ID} & \textbf{Text} & \textbf{Point} & \textbf{TSG} & \textbf{ICCD} & \textbf{IoU} & \textbf{mIoU} & \textbf{F1} & \textbf{Precision} & \textbf{Recall} \\
\midrule
A & \cmark & \xmark & \xmark & \xmark & 31.69 & 25.27 & 48.13 & \textbf{84.92} & 33.58 \\
B & \xmark & \cmark & \cmark & \cmark & \underline{35.95} & 44.23 & \underline{52.89} & 75.32 & 40.75 \\
\midrule
C & \cmark & \cmark & \xmark & \xmark & 27.53 & 32.29 & 43.18 & 30.90 & \underline{71.64} \\
D & \cmark & \cmark & \cmark & \xmark & 29.26 & \underline{46.69} & 45.27 & 31.86 & \textbf{78.17} \\
E & \cmark & \cmark & \xmark & \cmark & 27.66 & 38.18 & 43.33 & 32.13 & 66.53 \\
\rowcolor{rowours}
F & \cmark & \cmark & \cmark & \cmark & \textbf{59.44} & \textbf{59.29} & \textbf{74.56} & \underline{84.70} & 66.58 \\
\bottomrule
\end{tabular}%
}
\end{table*}

Rows A, B, and F establish the prompt-modality result. The full configuration improves mIoU by $34.02$ points over text-only SAM\,3 and by $15.06$ points over point-only decoding with identical TSG coordinates where points exist. Text supplies class semantics, while retrieved points supply target-domain location evidence. For 427 query--class pairs, or $19.56\%$ of the evaluation set, TSG produces no validated point. Row B assigns an empty mask to these pairs, whereas the joint configuration in row F uses the text-only fallback defined in Eq.~\eqref{eq:fallback}. The $15.06$-point difference is therefore an end-to-end prompt-mode result rather than a pure estimate of multimodal fusion restricted to point-bearing queries.

Rows C--F form the complete ICCD $\times$ TSG factorial with simultaneous text-and-point decoding. ICCD improves mIoU by $5.89$ points without TSG and $12.60$ points with TSG. The complete TSG selector improves mIoU by $14.40$ points with a raw bank and $21.11$ points with an ICCD bank. The resulting $6.71$-point factorial interaction is descriptive for this fixed support manifest and is not treated as a statistical interaction estimate across support seeds. Comparing rows D and F, ICCD also reduces aggregate pixel false positives from $558{,}582{,}430$ to $40{,}169{,}767$, a $92.81\%$ reduction. These results support complementary contributions from bank curation and the complete TSG point-selection procedure under the controlled five-shot setting.

TSG also reduces the mean number of positive points per query from $1{,}060.69$ to $21.00$ with the raw bank and from $883.68$ to $20.19$ with the ICCD bank. By applying similarity gating, connected-component filtering, peak extraction, and spatial NMS, TSG retains a compact set of higher-quality, spatially coherent prompts. This drastic reduction lowers the decoder's prompt-processing load and limits the opportunity for false-positive points to misguide mask formation. Because TSG changes both spatial organization and point count, the measured improvement represents the effect of the complete selector.

\subsection{Support Selection and Bank-Quality Evidence}
\label{sec:ablation_shots}

Because a one-versus-five-shot comparison changes both the number of observed appearances and the availability of cross-image reliability filtering, support-selection sensitivity and ICCD effectiveness are evaluated separately.

We first test whether the one- and five-shot results depend on the selected references while keeping the query set fixed.

\begin{table}[!t]
\centering
\caption{Repeated support sampling with fixed query sets. The mIoU values are percentages and are reported as mean $\pm$ sample standard deviation over seeds $17$, $29$, $43$, $71$, and $101$. The paired gain is five-shot minus one-shot in percentage points, and the final column is its 95\% Student-$t$ confidence interval.}
\label{tab:repeated_support_sampling}
\setlength{\tabcolsep}{4pt}
\renewcommand{\arraystretch}{1.12}
\begin{tabular}{lcccc}
\toprule
\textbf{Dataset} & \textbf{1-shot mIoU} & \textbf{5-shot mIoU} & \textbf{Paired gain} & \textbf{95\% CI} \\
\midrule
PC-59 & $66.01\pm0.42$ & $66.63\pm0.17$ & $+0.63\pm0.46$ & $[+0.05,+1.20]$ \\
AgML & $46.93\pm4.40$ & $58.51\pm1.30$ & $+11.59\pm3.58$ & $[+7.14,+16.03]$ \\
\bottomrule
\end{tabular}
\end{table}

Table~\ref{tab:repeated_support_sampling} shows a dataset-dependent effect. All five paired seeds give a positive five-shot gain. On PC-59, the mean gain is small and five shots mainly reduce support-selection variance. On AgML, one-shot performance varies more strongly with the selected reference, whereas five shots improve the mean by $11.59$ points and reduce the sample standard deviation from $4.40$ to $1.30$ points. In these experiments, moving from one to five references is more useful when class appearance is heterogeneous or domain-specific.

For prompt-level bank quality, rows D and F of Table~\ref{tab:ablation_components} provide the direct final-mask ICCD comparison. They hold supports, queries, TSG settings, and SAM\,3 mask generation fixed and change only whether the five-shot foreground bank is raw or ICCD-filtered.

\begin{table}[!t]
\centering
\caption{Effect of ICCD filtering on feature-bank size and spatial-prompt quality, averaged over five support selections. Feature-bank size is reported relative to the unfiltered raw bank ($100\%$). Random filtering retains the same number of descriptors as ICCD, separating the effect of descriptor selection from simple bank-size reduction. Prompt precision is the percentage of generated point prompts located within the target mask. False-positive points per query is the mean number located outside it. Arrows indicate the preferred direction, and bold marks the best prompt-quality value.}
\label{tab:iccd_prompt_quality}
\setlength{\tabcolsep}{6pt}
\renewcommand{\arraystretch}{1.12}
\resizebox{\textwidth}{!}{%
\begin{tabular}{lccc}
\toprule
\makecell{\textbf{Feature-bank}\\\textbf{construction}} & \makecell{\textbf{Feature-bank size}\\\textbf{(\% of raw)} $\downarrow$} & \makecell{\textbf{Prompt}\\\textbf{precision (\%)} $\uparrow$} & \makecell{\textbf{False-positive points}\\\textbf{per query} $\downarrow$} \\
\midrule
No filtering (raw bank) & $100.00$ & $68.88$ & $5.58$ \\
Random filtering (matched size) & $75.76$ & $71.83$ & $4.82$ \\
ICCD filtering (proposed) & $75.76$ & $\mathbf{87.98}$ & $\mathbf{1.72}$ \\
\bottomrule
\end{tabular}%
}
\end{table}

The prompt-level results in Table~\ref{tab:iccd_prompt_quality} show that the improvement is not explained by descriptor-count reduction alone. At the same retained bank size, ICCD gives substantially higher point precision and fewer false points than random filtering. Raw and ICCD banks have the same mean query hit rate of $90.94\%$, indicating that ICCD suppresses incorrect prompts without reducing the proportion of queries that receive at least one correct point. ICCD removes $24.24\%$ of effective descriptors on average and reduces cached matching-plus-TSG latency by $15.55\%$. These findings support a narrower claim than absolute necessity: a reference image can be useful without ICCD, but cross-image filtering substantially improves multi-reference bank quality by suppressing false-positive spatial evidence.

\subsection{ICCD Threshold Strategy}
\label{sec:ablation_iccd}

\begin{table}[!t]
\centering
\caption{Secondary ICCD-threshold sensitivity under the AgML largest-reference configuration used for this sweep. Because its reference setting differs from the controlled seed-17 five-shot component experiment, it is interpreted separately. All metrics are percentages, and bold marks the highest value in each column.}
\label{tab:ablation_iccd}
\setlength{\tabcolsep}{6pt}
\begin{tabular}{lcccc}
\toprule
\textbf{Threshold strategy} & \textbf{IoU} & \textbf{mIoU} & \textbf{F1} & \textbf{Precision} \\
\midrule
Fixed $\kappa = 0.65$      & 62.40 & 58.70 & 76.90 & 85.40 \\
Fixed $\kappa = 0.75$      & \textbf{62.90} & 59.50 & \textbf{77.20} & \textbf{85.70} \\
Fixed $\kappa = 0.82$      & 62.10 & \textbf{59.70} & 76.60 & 84.60 \\
Adaptive $Q_{75}$ (default) & 62.80 & 59.20 & \textbf{77.20}
                              & \textbf{85.70} \\
\bottomrule
\end{tabular}
\end{table}

The fixed and adaptive thresholds differ by at most $1.0$ mIoU percentage point. Fixed $\kappa = 0.75$ is marginally higher in IoU, while the adaptive threshold matches it in F1 and precision and comes within $0.1$ IoU point. The adaptive rule is retained as the proposed configuration because it avoids manual threshold selection when SegRAG is applied to a new domain: the per-class $Q_{75}$ statistic automatically scales the criterion to each category's observed coherence distribution.

\subsection{TSG Hyperparameter Sensitivity}
\label{sec:ablation_tsg}

\begin{table}[!t]
\centering
\caption{Secondary TSG sensitivity under the AgML largest-reference configuration used for this sweep. Because its reference setting differs from the controlled seed-17 five-shot component experiment, it is interpreted separately. All metrics are percentages. In each block, one parameter changes and the others remain at their defaults ($\tau_\ell=0.80$, $\eta_{\mathrm{cc}}=4$, $\delta=10$).}
\label{tab:ablation_tsg}
\setlength{\tabcolsep}{5pt}
\begin{tabular}{llccccc}
\toprule
\textbf{Parameter} & \textbf{Value} & \textbf{IoU} & \textbf{mIoU} & \textbf{F1} & \textbf{Precision} & \textbf{Recall} \\
\midrule
\multirow{3}{*}{$\tau_\ell$}
  & $0.75$                  & 66.30 & 58.70 & 79.70 & 84.90 & 75.20 \\
  & $0.80$ \emph{(default)} & 62.80 & 59.20 & 77.20 & 85.70 & 70.20 \\
  & $0.85$                  & 51.20 & 57.10 & 67.70 & 83.80 & 56.80 \\
\midrule
\multirow{3}{*}{$\eta_{\mathrm{cc}}$}
  & $1$                     & 60.50 & 56.80 & 75.40 & 80.50 & 70.80 \\
  & $4$ \emph{(default)}    & 62.80 & 59.20 & 77.20 & 85.70 & 70.20 \\
  & $8$                     & 62.20 & 59.90 & 76.70 & 86.00 & 69.20 \\
\midrule
\multirow{3}{*}{$\delta$}
  & $5$                     & 61.40 & 59.50 & 76.10 & 82.30 & 70.80 \\
  & $10$ \emph{(default)}   & 62.80 & 59.20 & 77.20 & 85.70 & 70.20 \\
  & $15$                    & 62.70 & 59.00 & 77.10 & 85.70 & 70.10 \\
\bottomrule
\end{tabular}
\end{table}

The similarity threshold has the largest effect within the tested ranges. A false-positive point can strongly steer the SAM\,3 decoder toward an incorrect spatial region, whereas a missed point leaves the text-only path available. Within the $\tau_\ell$ sweep, the default $\tau_\ell=0.80$ gives the highest mIoU and precision. $\tau_\ell=0.75$ is preferable when recall is more important, while $\tau_\ell=0.85$ discards too many valid candidates and reduces recall to $56.8\%$. The component size $\eta_{\mathrm{cc}}$ and NMS distance $\delta$ are less sensitive once singleton noise is excluded, although the best operating point depends on the relative cost of false positives and missed regions.

\subsection{DINOv2 versus DINOv3}
\label{sec:ablation_backbone}

\begin{table}[!t]
\centering
\caption{Secondary feature-extractor ablation under the AgML largest-reference configuration used for this comparison. Both rows use the same SegRAG pipeline and SAM\,3 decoder. Because the reference setting differs from the controlled seed-17 five-shot component experiment, the results are interpreted separately. All values are percentages. DINOv3 is the default backbone.}
\label{tab:ablation_backbone}
\setlength{\tabcolsep}{5pt}
\resizebox{\linewidth}{!}{%
\begin{tabular}{lccccc}
\toprule
\textbf{Feature extractor} & \textbf{IoU} & \textbf{mIoU} & \textbf{F1}
  & \textbf{Precision} & \textbf{Recall} \\
\midrule
DINOv2~\cite{oquab2024dinov2} & $\mathbf{65.07}$ & $\mathbf{59.66}$ & $\mathbf{78.84}$ & $76.39$ & $\mathbf{81.45}$ \\
DINOv3~\cite{simeoni2025dinov3} \emph{(default)} & $62.84$ & $59.24$ & $77.18$ & $\mathbf{85.71}$ & $70.20$ \\
\bottomrule
\end{tabular}%
}
\end{table}

The two encoders present a precision--recall trade-off: DINOv2 recovers more foreground pixels ($81.45$ vs $70.20$ recall), whereas DINOv3 produces fewer false positives ($85.71$ vs $76.39$ precision). Their mIoU values differ by only $0.42$ points, so the AgML result cannot be attributed solely to DINOv3 descriptors. This experiment keeps the SAM\,3 decoder fixed and therefore does not isolate dependence on SAM\,3 relative to earlier or alternative promptable decoders. DINOv3 remains the default because its higher precision reduces false-positive point prompts. DINOv2 may be preferable when foreground recall is the primary objective.

Taken together, the controlled component ablation shows complementary contributions from bank curation, the complete TSG selector, and joint text-and-point prompting. The secondary sensitivity studies show that the size and precision--recall balance of these effects depend on the operating point and feature encoder. These results also delimit the claim: SegRAG remains dependent on the capabilities of its frozen encoder and decoder, ICCD is a multi-reference false-positive suppression mechanism rather than a prerequisite for all retrieval gains, and the present TSG comparison does not separate spatial topology from prompt-count reduction.

\section{Discussion and Implications}
\label{sec:discussion_implications}

\subsection{What the empirical evidence reveals}

The experiments distinguish knowing a class name from grounding that name under the conditions in which an image was acquired. Text-only SAM\,3 already provides a strong result when the requested class resembles its broadly learned concept. Adding retrieved points gives smaller, consistent gains on the standard benchmarks, but becomes more consequential on AgML, where several crops appear from unfamiliar viewpoints or at growth stages likely underrepresented in generic visual pretraining data. This contrast suggests that external visual memory is not uniformly beneficial. Its value increases under domain shift, when semantic intent is available but the pretrained model lacks spatial evidence for the target appearance.

The support-sampling results refine this interpretation. On PC-59, moving from one to five references mainly reduces sensitivity to which exemplar is selected. On AgML, it also improves the mean because one image rarely represents the observed appearance variation. The controlled component experiment identifies two different filtering effects. Raw descriptors recover foreground broadly but also produce many false prompts, whereas ICCD sacrifices some recall to obtain a much larger increase in precision. ICCD should therefore be understood as a multi-reference evidence-quality mechanism, not as the source of retrieval itself. The complete TSG selector turns hundreds of dense matches per query into roughly 20 point prompts while improving mIoU with both raw and ICCD banks. Because this operation changes both spatial organization and prompt count, the experiment supports the selector as implemented rather than topology alone. The prompt-modality comparison provides a third distinction: class text supplies semantic identity, while retrieved points supply target-domain location evidence. Their measured difference is an end-to-end result that also includes zero-point queries. The DINOv2 comparison further shows that the outcome is not explained solely by adopting DINOv3: the encoders reach similar mIoU through different precision--recall trade-offs.

\subsection{Position relative to existing approaches}

The controlled comparison with SAM\,3 text-only isolates the central intervention: the decoder family and class text remain fixed, while SegRAG adds spatial evidence retrieved from annotated references. Comparisons with Matcher, IPSeg, GF-SAM, and FSSDINO~\cite{liu2024matcher,tang2024ipseg,zhang2024gfsam,zakir2026fssdino} address a broader question. These methods also exploit visual supports, but do so through their own episodic matching or prompt-generation designs and different backbone--decoder combinations. Their scores establish the competitiveness of SegRAG under matched support and query protocols; they do not isolate a single architectural cause.

No Time to Train! is the closest comparison in its use of a descriptor memory~\cite{espinosa2025notimetotrain}, but the location of retrieval in the processing chain differs. It labels masks after automatic proposal generation, whereas SegRAG retrieves before decoding so that the evidence can influence which region is segmented. The observed score difference cannot be attributed to this ordering alone because the methods use different DINO and SAM generations. Synthetic-prototype methods make another trade-off: FOSSIL and FreeDA~\cite{barsellotti2024fossil,barsellotti2024freeda} avoid annotated target supports, while SegRAG accepts that annotation cost in exchange for evidence drawn from the deployment domain. The Sa2VA result~\cite{sa2va} provides a useful boundary rather than another universal ranking. Sa2VA is stronger on PC-59 without supports, whereas SegRAG is stronger on the evaluated agricultural transfer setting when annotated domain evidence is available. Together, these comparisons place SegRAG as a specialized adaptation mechanism for settings where a small visual memory can be maintained, not as a replacement for support-free models in every regime.

\subsection{Information-management and practical implications}

SegRAG frames segmentation adaptation as the organization and retrieval of external visual evidence rather than as the modification of model parameters. Annotated references form a class-indexed resource, allowing visual evidence for a class to be added, inspected, corrected, and updated independently of the frozen encoder and decoder. Retrieval converts this maintained evidence into mask-generation prompts before decoding, allowing it to influence which regions are segmented. Evidence acquisition, quality control, retrieval, and use therefore remain explicit and independently manageable stages.

This design is most practical when class definitions change after deployment, only limited annotations are available, or repeated fine-tuning is difficult to validate and maintain. The approach still requires annotated reference masks and computational resources. Feature banks incur offline construction and storage costs, and query matching adds latency relative to text-only inference. The appropriate operational comparison is therefore between the cost of maintaining an external memory and the cost of retraining, deploying a larger general-purpose model, or accepting failures under domain shift. Which option is preferable depends on annotation availability, update frequency, hardware constraints, and the consequence of false prompts.

\subsection{Scope, robustness, and external validity}

The experiments establish SegRAG for two-dimensional RGB segmentation with English class names across four established benchmarks and the selected AgML collections. The AgML results demonstrate effective transfer to an agricultural domain whose visual characteristics differ from general-purpose imagery. This indicates potential for other specialized domains, although separate evaluation remains necessary. Medical imaging, industrial inspection, remote sensing, biodiversity monitoring, video, and non-RGB sensing are promising settings for this separate validation.

Robustness depends first on the visual memory. Incorrect labels, incomplete masks, limited subtype coverage, or a bank dominated by one acquisition condition can all produce misleading points. The current class-keyed retrieval also loads the complete retained bank without first adapting it to query metadata such as sensor, illumination, scale, or viewpoint. Context-aware reference selection based on metadata or coarse query-to-reference matching could reduce this noise and should be evaluated independently. At the model level, SegRAG assumes that DINO patch descriptors preserve useful local correspondence and that SAM\,3 responds consistently to simultaneous text and point prompts. Results may change with other encoders, decoders, languages, or resource-constrained platforms.

Retrieved points provide partial interpretability by showing which class memory generated a spatial cue and which references supplied the corresponding evidence. They do not explain every decision made by SAM\,3 when converting those cues into a pixel-level mask. Future work can extend this traceability through automatic reference-quality monitoring, multilingual prompt evaluation, and acquisition-aware memory selection.

\section{Conclusion}
\label{sec:conclusion}

This paper introduced SegRAG, a retrieval-augmented inference-time pipeline for open-vocabulary semantic segmentation. SegRAG uses annotated segmentation examples to build a feature bank for each class. ICCD curates the bank by retaining descriptors that are reliable across the reference images, thereby reducing false-positive point prompts, while TSG matches the curated features to the query image and converts spatially coherent matches into a compact set of point prompts. During inference, SegRAG passes the retrieved points and class text together to the frozen SAM\,3 model in a single call to generate the segmentation mask without updating model weights. Experiments across standard segmentation benchmarks and agricultural imagery showed that SegRAG outperformed the evaluated matched reference-based baselines and consistently improved over SAM\,3 text-only prompting, with its strongest benefit appearing under domain shift, where it localized categories missed by text-only grounding. SegRAG formulates domain adaptation for segmentation as an information organization and retrieval problem, with annotated visual knowledge maintained as an external, class-indexed resource that can be filtered, retrieved, inspected, and updated independently of the segmentation model. This provides a practical mechanism for reusing domain-specific visual knowledge to adapt frozen segmentation models when text alone does not provide sufficient spatial evidence.

\section*{Acknowledgements}
This research was supported by KU Research Center for Advanced Intelligent Systems (AIS), Khalifa University of Science and Technology (KU-AIS).

\section*{Author Contributions}
Abderrahmene Boudiaf: Conceptualization, Methodology, Software, Formal analysis, Investigation, Writing -- original draft, Writing -- review and editing. Irfan Hussain: Supervision, Funding acquisition. Sajid Javed: Supervision, Writing -- review and editing, Validation.

\section*{Declaration of Competing Interest}
The authors declare that they have no known competing financial interests or personal relationships that could have appeared to influence the work reported in this paper.

\section*{Data Availability}
The datasets used in this study are publicly available. Evaluations on standard benchmarks were conducted on ADE20K-150, Cityscapes, PC-59, and LVIS, all of which are accessible through their official sources. Agricultural segmentation experiments were conducted on datasets from the AgML benchmark, available at \url{https://github.com/Project-AgML/AgML}. The released SegRAG repository includes the AgML support and query manifests under \texttt{splits/agml/}, including image identifiers, relative file names, image sizes, and annotation identifiers. The code implementing SegRAG, including the ICCD and TSG pipelines, is available at \url{https://github.com/boudiafA/SegRAG}.

\bibliographystyle{elsarticle-num}
\bibliography{references2}

\end{document}